\documentclass[10pt,twocolumn,letterpaper]{article}

\usepackage[pagenumbers]{iccv} 
\usepackage{algorithm}
\usepackage{algorithmic}
\usepackage{array}
\definecolor{iccvblue}{rgb}{0.21,0.49,0.74}
\usepackage{xcolor}
\usepackage[pagebackref,breaklinks,colorlinks,allcolors=iccvblue]{hyperref}

\NewDocumentCommand{\kaiyan}
{ mO{} }{\textcolor{purple}{\textsuperscript{\textit{kaiyan}}\textsf{\textbf{\small[#1]}}}}
\def\paperID{11028} 
\def\confName{ICCV}
\def\confYear{2025}

\title{Video-T1: Test-Time Scaling for Video Generation}

\author{Fangfu Liu$^{1}$\footnotemark[1], Hanyang Wang$^{1}$\footnotemark[1], Yimo Cai$^{1}$, Kaiyan Zhang$^{1}$, Xiaohang Zhan$^{2}$, Yueqi Duan$^{1}$\footnotemark[2]\\
$^{1}$Tsinghua University, 
$^{2}$Tencent}

\begin{document}
\twocolumn[{
 \renewcommand\twocolumn[1][]{#1}
\maketitle
 \thispagestyle{empty}
 \pagestyle{empty}
 \begin{center}
     \captionsetup{type=figure}
    \vspace{-2em} \includegraphics[width=1\linewidth]{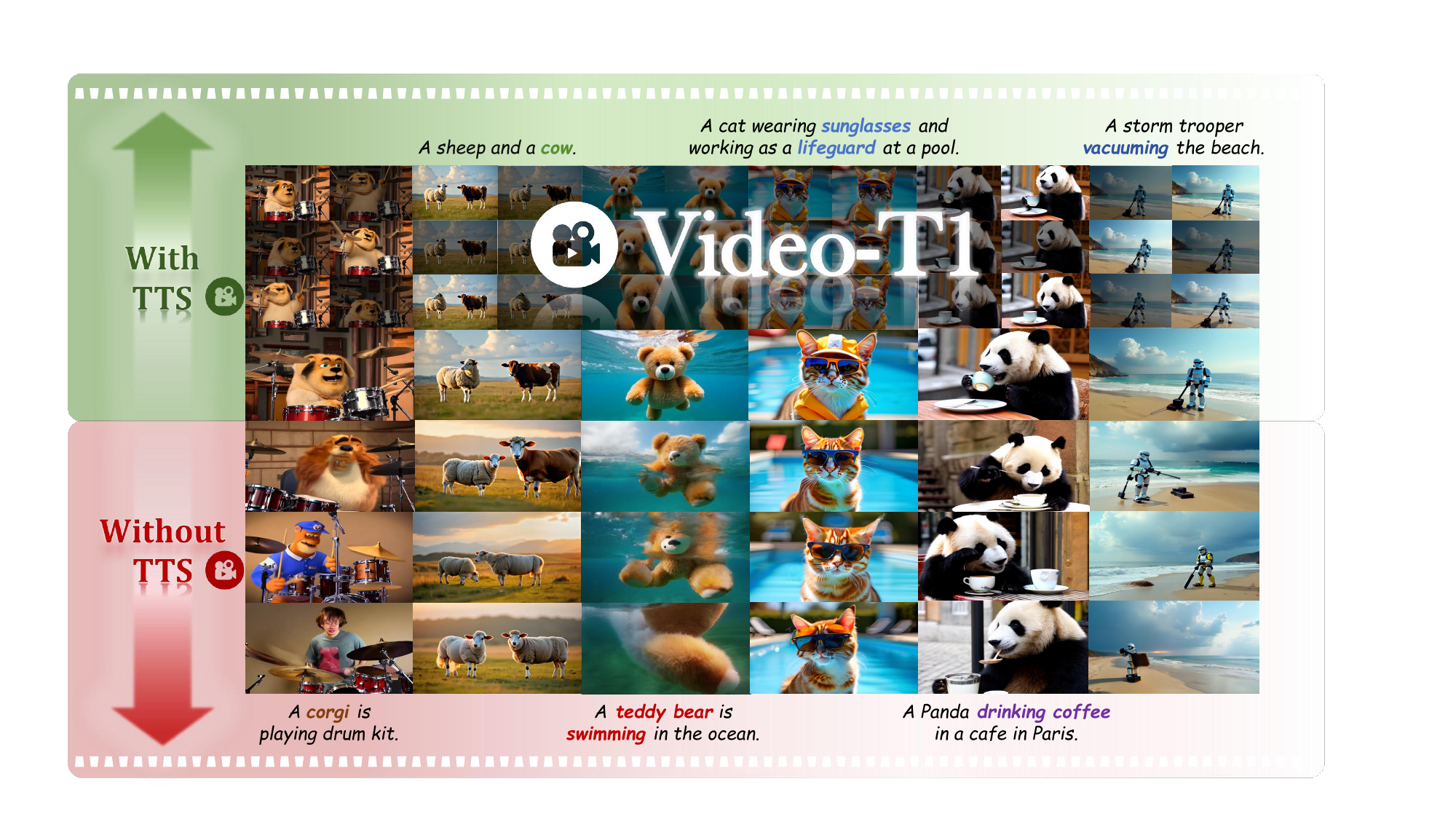}
    \vspace{-2em}
     \captionof{figure}{{\textbf{Video-T1:} We present the generative effects and performance improvements of video generation under Test-Time Scaling (TTS) settings. The videos generated with TTS are of higher quality and more consistent with the prompt than those generated without TTS.}}
     \label{fig:overview}
     \vspace{0.1cm}
 \end{center}
}]

\maketitle
\renewcommand{\thefootnote}{\fnsymbol{footnote}}
\footnotetext[1]{Equal contribution. $\dagger$ The corresponding author.}
\renewcommand{\thefootnote}{\arabic{footnote}}
\begin{abstract}
With the scale capability of increasing training data, model size, and computational cost, video generation has achieved impressive results in digital creation, enabling users to express creativity across various domains. Recently, researchers in Large Language Models (LLMs) have expanded the scaling to test-time, which can significantly improve LLM performance by using more inference-time computation. Instead of scaling up video foundation models through expensive training costs, we explore the power of Test-Time Scaling (TTS) in video generation, aiming to answer the question: if a video generation model is allowed to use non-trivial amount of inference-time compute, how much can it improve generation quality given a challenging text prompt. In this work, we reinterpret the test-time scaling of video generation as a searching problem to sample better trajectories from Gaussian noise space to the target video distribution. Specifically, we build the search space with test-time verifiers to provide feedback and heuristic algorithms to guide searching process. Given a text prompt, we first explore an intuitive linear search strategy by increasing noise candidates at inference time. As full-step denoising all frames simultaneously requires heavy test-time computation costs, we further design a more efficient TTS method for video generation called Tree-of-Frames (ToF) that adaptively expands and prunes video branches in an autoregressive manner. Extensive experiments on text-conditioned video generation benchmarks demonstrate that increasing test-time compute consistently leads to significant improvements in the quality of videos. Project Page: \url{https://liuff19.github.io/Video-T1}.
\end{abstract}    

\section{Introduction}

\label{sec:intro}
The field of generative modeling has witnessed remarkable progress in recent years~\cite{rombach2022high-stable-diffusion, saharia2022photorealistic-imagen, achiam2023gpt, yang2024cogvideox}, with applications spanning from image and text generation to more complex tasks, such as video synthesis. Among these, video generation~\cite{kong2024hunyuanvideo, li2024survey-video-gen} stands out due to its potential to revolutionize digital content creation, enabling the automatic production of high-quality videos from simple textual descriptions~\cite{zeng2024dawn}. This capability has profound implications for various industries~\cite{liu2024video-p2p, li2024survey-video-gen, kong2024hunyuanvideo} (\eg, entertainment, education, and advertisements). The pivotal factor of the exponential growth in video generation lies in the scaling-up capability by training with an expanding volume of data, more computational sources, and larger model sizes~\cite{peebles2023scalable-DiT, kong2024hunyuanvideo}. This scaling behavior during the training process, commonly referred to as \textit{Scaling Laws}~\cite{hoffmann2022training-compute-optimal, kaplan2020scaling-law, ruan2025observational-scaling, peebles2023scalable-DiT}, plays a crucial guiding role in the advancement of generative models with progressively higher capabilities.

Despite these advancements, generating high-quality videos remains challenging due to the need for maintaining temporal coherence and capturing complex dynamics across frames~\cite{zeng2024dawn}. While scaling video generation methods in the training process~\cite{ma2025step-t2v, kong2024hunyuanvideo} has yielded significant improvements, it is inherently limited by high costs and resource demands, making it challenging to scale further.
Recently, researchers in LLMs have expanded the study of scaling to the test-time~\cite{liu2025can-TTS} (\eg, DeepSeek-R1~\cite{guo2025deepseek} and OpenAI o1~\cite{jaech2024openai-o1}) and demonstrated that Test-time Scaling (TTS) can significantly improve the performance of LLMs with more contextually appropriate responses by allocating additional computation at inference time~\cite{team2025kimi, wang2025thoughts-QW, guo2025deepseek, jaech2024openai-o1}.

In this paper, we propose to investigate Test-Time Scaling (TTS) for video generation. Specifically, we aim to answer the question: \textit{If a video generation model is permitted to use the larger amount of inference-time computation, how much can it improve the generation quality for challenging text prompts?} We seek to explore the potential of TTS to enhance video generation without the need for expensive retraining or model enlargement. 
To understand the benefits of scaling up test-time computation in video diffusion, we propose a general framework for TTS video generation, called \textbf{Video-T1}, which reinterprets the TTS of video generation as a searching problem within the space of possible video trajectories originating from Gaussian noise. The key insight is to scale the search space at test time with increased computation so that we can find a broader range of potential solutions to generate higher-quality and text-aligned videos. In our search framework, we introduce test-time verifiers to assess the quality of intermediate results and heuristic algorithms to navigate the search space efficiently. Initially, we conduct a straightforward random linear search strategy by sampling N noise candidates in parallel and selecting the one that scores the highest per a test-time verifier. However, recognizing the computational intensity of this approach, particularly when denoising all video frames simultaneously, we introduce a more efficient framework called Tree-of-Frames (ToF). ToF operates in an autoregressive manner under a tree structure, which leverages the feedback from verifiers and adaptively expands and prunes branches of video frames to balance computational cost and generation quality. Through extensive experiment on text-conditioned video generation benchmark, our findings reveal that increasing test-time compute leads to substantial improvements in the quality and human-preference alignment of samples generated by video generation models. Longer term, this offers a significant promise on how to leverage inference-time computation to achieve superior results in computer vision. (See qualitative results gallery in Figure~\ref{fig:overview} and
quantitative results in Figure~\ref{fig:fig0}). Our contributions are summarized as follows:
\begin{figure*}[t]
    \centering
    \includegraphics[width=\linewidth]{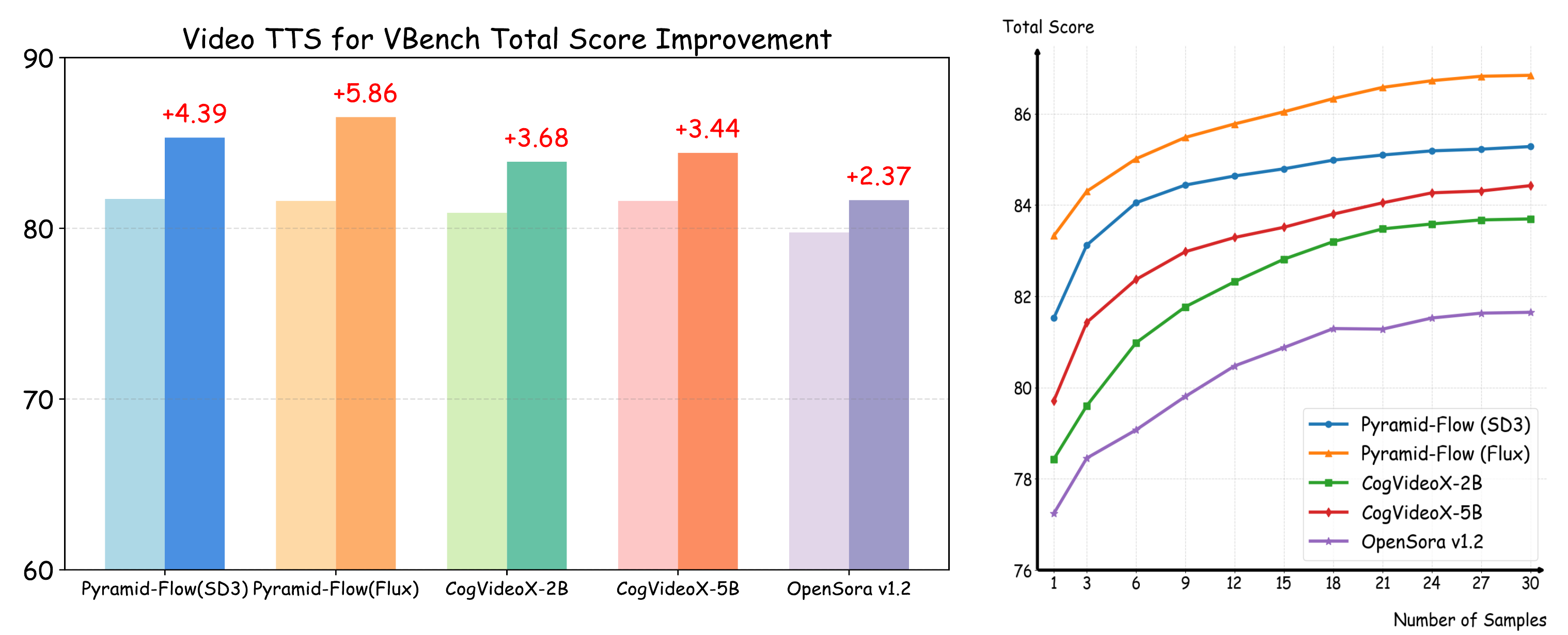}
    \vspace{-2em}
    \caption{\textbf{Results of Test-Time Scaling for Video Generation.} As the number of samples in the search space increases by scaling test-time computation (TTS), the models' performance exhibits consistent improvement (In the bar chart, light colors correspond to the results without TTS, while dark colors represent the improvement after TTS.).}
    \label{fig:fig0}
    \vspace{-1em}
\end{figure*}
\begin{itemize}
    \item We propose a fundamental framework \textit{Video-T1} for test-time scaling for video generation, which reinterprets this process as a search problem to sample better video trajectories. We show that scaling the search space of video generation can boost video performance across different dimensions of the benchmark.
    \item We carefully build the search space in test-time scaling by test-time verifiers to provide feedback and heuristic algorithms (\ie, a straightforward random linear search and ToF search for more efficient test-time scaling) to guide the search process. 
    \item Extensive experiments demonstrate that scaling the search space of video generation can boost the performance of various video generation models across different dimensions of the benchmark, and our proposed ToF search can significantly reduce scaling cost when achieving high-quality results.
\end{itemize}
\section{Related Work}
\label{sec:related work}
\noindent \textbf{Test-Time Scaling in LLMs.} Recent advancements have demonstrated the effectiveness of test-time scaling (TTS) methods such as chain-of-thought prompting~\cite{wei2022chain,madaan2023self}, outcome reward models, and process reward models~\citep{lightman2023let,wang2023math,zhangopenprm} in enhancing the reasoning capabilities of large language models (LLMs) during inference stages. Notable examples include implementations in OpenAI o1~\cite{jaech2024openai-o1} and DeepSeek-R1~\cite{guo2025deepseek}. These methods promote the generation of intermediate reasoning steps, resulting in more precise responses.
These researches suggests that reallocating computational resources from pre-training~\citep{kaplan2020scaling} to test-time can enhance performance more efficiently~\citep{snell2024scaling,liu2025can-TTS}.
Moreover, strategies like self-consistency~\citep{wang2022self,chen2023universal}, best-of-N~\citep{stiennon2020learning,nakano2021webgpt}, Monte Carlo Tree Search~\citep{zhou2023language,xie2024monte}, and Reward-guided Search~\citep{deng2023reward, khanov2024args} employ diverse generation techniques and sophisticated aggregation methods, often facilitated by process reward models. These approaches help in producing diverse and integrated outputs.
Additionally, DeepSeek R1~\cite{guo2025deepseek} utilizes outcome-based reinforcement learning techniques, like group relative policy optimization~\citep{shao2024deepseekmath}, to enhance the reasoning capabilities of pre-trained models. The combination of parallel and sequential generation techniques in these models represents a nuanced approach to generating contextually appropriate outputs, thereby establishing new operational standards for LLMs in complex problem-solving scenarios.

\noindent \textbf{Test-Time Scaling in Computer Vision.} In both the visual understanding and visual generation fields, researchers have investigated various test-time scaling methods to further push the performance boundaries. With the success of test-time scaling methods in LLMs, several recent vision language models (VLMs) ~\cite{xu2025llavacotletvisionlanguage, thawakar2025llamavo1rethinkingstepbystepvisual}  utilized step-by-step reasoning capability enhanced by test-time scaling methods and surpassed larger models in visual question-answering tasks. Recent investigations on image diffusion models have demonstrated that image diffusion models' generation quality could be further enhanced with test-time scaling methods~\cite{guo2025image-cot}. With verifiers providing judgments and algorithms selecting better candidates, image diffusion models consistently improve their performance across generation tasks by scaling up inference time~\cite{ma2025inferencetimescalingdiffusionmodels}.

\noindent \textbf{Video Generation.} Efficient and high-quality video generation has attracted increasing attention due to its wide applications in areas~\cite{liu2024physics3d,liu2024reconx,sun2024dimensionxcreate3d4d,ling2024motionclone}. With the success of diffusion models~\cite{ho2020ddpm, rombach2022ldm} in text-to-image generation, several studies have extended them to text-to-video (T2V) tasks, achieving promising results. One line of work~\cite{blattmann2023svd,liu2024sora,chen2024videocrafter2,hong2022cogvideo,yang2024cogvideox} improves video quality by scaling up diffusion transformer (DiT)~\cite{peebles2023DiT} pre-training, leading to high visual fidelity and smoother motion. These models have reached near-production-level performance but require extensive computational resources, especially for long videos~\cite{jin2024pyramid}.
Another line of work~\cite{chen2025diffusion_forcing,xu2022poisson,kondratyuk2024videopoet,jin2024video-lavit,deng2024NOVA,jin2024pyramid} combines diffusion models with autoregressive mechanisms to better handle long and complex videos. For example, NOVA~\cite{deng2024NOVA} generates videos by predicting frames sequentially over time while sampling tokens in random spatial order, unifying various generation tasks into a single framework. Pyramid-Flow~\cite{jin2024pyramid} redefines the generation process as a multi-scale trajectory over compressed representations, using spatial and temporal pyramids to reduce training costs while maintaining quality. The autoregressive approaches show strong potential to generate longer, coherent, and high-quality videos with improved efficiency, making them a promising direction for future research.
\begin{figure*}[t]
    \centering
    \includegraphics[width=\textwidth]{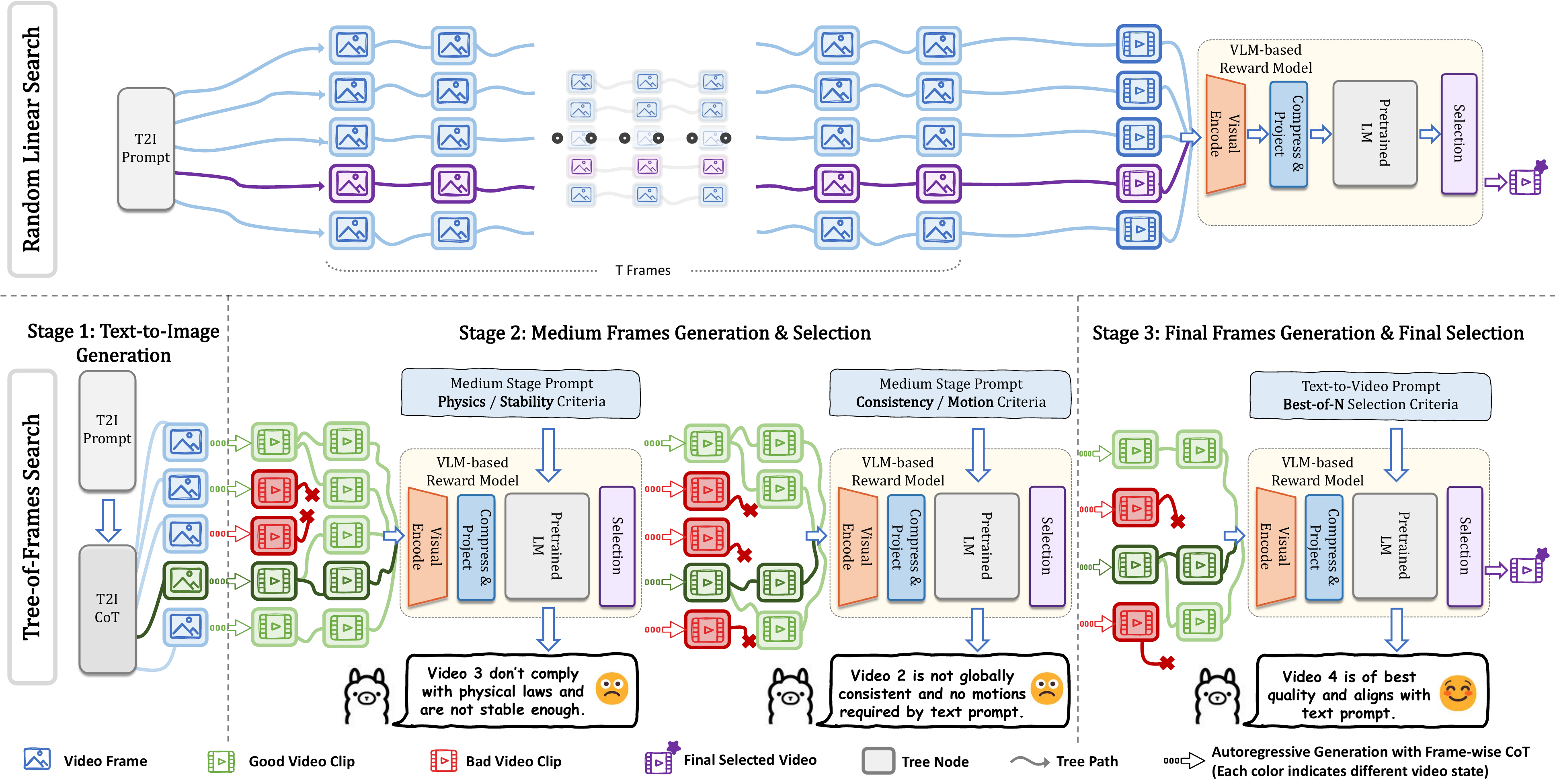}
    \caption{\textbf{Pipeline of Test-Time Scaling for Video Generation.} \textit{Top:} \textbf{Random Linear Search} for TTS video generation is to randomly sample Gaussian noises, prompt the video generator to generate sequential of video clips through step-by-step denoising in a linear manner, and select the highest score form the test verifiers. \textit{Bottom:} \textbf{Tree of Frames (ToF) Search} for TTS video generation is to divide the video generation process into three stages: (a) the first stage performs image-level alignment that influences the later frames; (b) the second stage is to apply dynamic prompt in test verifiers $\mathcal{V}$ to focus on motion stability, physical plausibility to provide feedback that guides heuristic searching process; (c) the last stage assesses the overall quality of the video and select the video with highest alignment with text prompts.  }
    \label{fig:pipeline}
\end{figure*}

\section{Method}
\subsection{How to Scale Video Generation at Test Time}
In the realm of LLMs, researchers have explored the benefits of scaling up test-time computation to boost model performance. Several key factors have been identified that shape the effectiveness of test-time scaling strategies in LLMs, such as the choice of policy models, process reward models (PRMs), and varying levels of problem difficulty~\cite{liu2025can-TTS, snell2024scaling-optim}. 
Similarly, Test-Time Scaling (TTS) in video generation hinges on key components like different video generation models, multimodal evaluation models, and the complexity of prompts across diverse benchmark dimensions. However, unlike LLMs, video generation poses specific challenges. First, videos inherently exhibit strong temporal continuity, meaning that while they consist of discrete frames, ensuring smooth transitions between frames is essential for perceptually coherent results. Second, state-of-the-art video generation models are primarily based on diffusion models, which employ a multi-step denoising process that complicates the direct scaling of computational resources. 
These factors introduce additional complexities: test-time scaling in video generation must simultaneously address both spatial (frame-level) quality and temporal consistency while also considering the heavy iterative diffusion denoising process.

To address these challenges, we propose to reinterpret video TTS as a path-search problem to sample better trajectories from pure Gaussian noise space to the target video distribution. The key insight is to scale the search space at test time with increased computation so that we can explore a broader range of potential solutions to generate higher-quality and text-aligned videos. Taking a closer look at this scheme, a video can be represented as a sequence of discrete frames. Considering the temporal nature of the frame sequence, it can be modeled as a chain-like architecture, where the video generation resembles the growth of a degenerate tree (\ie, a tree where each non-leaf node has exactly one child -- rooted in the Gaussian noise space of the video domain). In this way, we formalize the generation of a high-quality video as a searching problem: starting from an initial root node, we seek a path through $T$ steps that reaches a leaf node, maximizing the quality along the generated sequence. To build such a search space, we define several key components:
\begin{itemize}
\item \textbf{Video Generator} $\mathbf{\mathcal{G}}$: Video generation models, which generate videos from given text prompts by the multi-step denoising process. Formally, we define:
\begin{equation} 
\mathcal{G}: c \to \mathbb{R}^{H \times W \times C \times T}, \end{equation} where $c$ represents the input text condition, and the output is a generated video with $T$ frames.

\item \textbf{Test Verifiers} $\mathcal{V}$: Multimodal evaluation models that assess the quality of generated videos and assign a final score to provide feedback in the generation process. This can be expressed as:
\begin{equation} 
\mathcal{V}: \mathbb{R}^{H \times W \times C \times T} \times c \to \mathbb{R}, 
\end{equation} 
where the function takes both the generated video and the input condition to produce a scalar quality score.

\item \textbf{Heuristic Search Algorithms} $f$: The optimization methods that leverage feedback from the verifier to guide the search trajectory, ultimately finding better video sequences. We define this as:
\begin{equation} 
f: \mathcal{G} \times \mathcal{V} \times (\mathbb{R}^{H \times W \times C})^N \times c \to \mathbb{R}^{H \times W \times C \times T}, \end{equation}
where $(\mathbb{R}^{H \times W \times C})^N$ represents the set of 
$N$ initial noise samples (\ie, root nodes in the search forest), and $\mathbb{R}^{H \times W \times C \times T}$ denotes the final selected video sequence (\ie, a path from a root node to a leaf node at depth $T$).
\end{itemize}

\subsection{Random Linear Search}

A straightforward approach for TTS video generation is to randomly sample Gaussian noises, prompt $\mathcal{G}$ to generate complete video sequences by performing the full denoising process for each sample, and perform the Best-of-N selection to obtain the one with the highest score from the test verifiers $\mathcal{V}$. We refer to this method as \textbf{random linear search} (the top of Figure~\ref{fig:pipeline}), as it performs step-by-step denoising in a linear manner along the noise dimension. In this search algorithm, each noise sample deterministically corresponds to a determined video output, and the only scaling factor for test-time scaling is the number of noise samples $N$, leading to a computational cost that increases linearly with the number of samples.

From a more structural perspective, random linear search can be interpreted as a forest consisting of $N$ degenerate trees, where each tree represents an independent sequence of $T$ denoising steps. The search task then reduces to selecting a better length-$T$ path among them. The total number of nodes in the forest is $TN$, leading to a generation time complexity of $
O(TN)$. Since each video evaluation requires a constant-time assessment of its quality, the evaluation cost per sample is $O(1)$, resulting in an overall quadratic time and space complexity $O(TN)$. 

While random linear search provides a simple baseline, its linear structure introduces two inherent limitations: 1) \textbf{Simplicity of linear structure.} Although the final path selects a single branch, the tight bounds of this approach require exhaustive traversal of the entire space, lacking efficient optimization mechanisms. 2) \textbf{Isolation of independent structure}. Without any feedback or interaction mechanisms between trees, it introduces additional randomness, making it slower for test-time scaling.

\begin{algorithm}[ht]
\caption{Random Linear Search}
\label{alg:linear_search}
\begin{algorithmic}[1]
\REQUIRE Number of noise samples $N$, video frame length $T$, verifier $\mathcal{V}$, video generator $\mathcal{G}$ and decoder $\mathcal{D}$, Gaussian noise distribution $\mathcal{N}$
\ENSURE Video $\hat{v}$ with the highest verifier score
\vspace{2mm}
\STATE Initialize empty set $\mathcal{C} \gets \{\}$
\FOR{$i = 1$ to $N$}
    \STATE Sample initial noise $z^{(i)} \sim \mathcal{N}$
    \STATE Initialize $x^{(i)}_0 \gets z^{(i)}$
    \FOR{$t = 1$ to $T-1$}
        \STATE $\{x^{(i)}_{\tau}\mid \tau \in [0,t]\} \gets \mathcal{G}(\{x^{(i)}_{\tau}\mid \tau \in [0,t-1]\}, t)$
    \ENDFOR
    \STATE Decode video $v^{(i)} \gets \mathcal{D}(\{x^{(i)}_{\tau}\mid \tau \in [0,T-1]\})$
    \STATE Compute score $s^{(i)} \gets \mathcal{V}(v^{(i)})$
    \STATE Add $(v^{(i)}, s^{(i)})$ to $\mathcal{C}$
\ENDFOR
\STATE Final verify $\hat{v} \gets \arg\max\limits_{(v, s) \in \mathcal{C}} s$
\RETURN $\hat{v}$
\end{algorithmic}
\end{algorithm}

\subsection{Tree-of-Frames Search}
Random linear search is essentially adopted by a Best-of-N strategy that scales test-time computation through increasing the number of initial noise samples $N$. However, this approach requires a fixed time complexity of $O(TN)$ as analyzed above, which becomes increasingly inefficient as either $N$ scales up or the video length $T$ grows, making it impractical for long video generation or high-quality sampling at larger scales. To address this limitation and achieve a better balance between video quality and test-time computational efficiency, we propose \textbf{Tree-of-Frames (ToF) Search} (Algorithm~\ref{alg:tof_search}), which leverages the sequential generation capability of autoregressive models (unlike diffusion models that denoise the entire video sequence simultaneously), introducing inference-time reasoning along the temporal dimension. This approach enables a more flexible and scalable video generation process, structured into three distinct stages: given a text prompt as input, (a) the first stage is to generate the initial frame with text-alignment on various dimensions (\eg, spatial relation, appearance style, color), which strongly impacts later frames due the continuity of video frames; (b) the second stage focuses on generating intermediate frames which should consider the key factors like subject consistency, motion stability, even physics plausibility to guarantee a smooth video flow; (c) the final stage is dedicated to assessing the overall video quality and alignment with text prompts. According to the goal of three stages, we meticulously design three key techniques in ToF search algorithm: \emph{image-level alignment}, \emph{hierarchical prompting}, and \emph{heuristic pruning}.

\noindent \textbf{Image-level alignment.}
Different from LLMs, video generation involves both spatial and temporal dimensions. Along the spatial axis, video frames are generated through step-wise denoising employed in diffusion models. Inspired by the Chain-of-Thought (CoT) reasoning mechanism in image generation~\cite{guo2025image-cot}, we introduce a progressive evaluation strategy at the frame level to dynamically scale computation during the denoising process. Specifically, during the denoising of each frame, a potential test verifier evaluates whether the partially denoised image has reached sufficient clarity for reliable assessment. Early stage frames often remain too blurry for a meaningful evaluation, which could mislead the scoring of frame quality. Once the frame reaches a visually informative state, the model further assesses its potential to evolve into a high-quality final image. By performing early rejection of low-potential candidates and allocating compute toward promising trajectories, image-level scaling ensures more efficient use of resources during inference.

\noindent \textbf{Hierarchical prompting.}
From a spatial perspective, each video frame is generated as an independent image. However, different frames play distinct roles in shaping the video's narrative and temporal coherence. With the analysis above, we design a hierarchical prompting strategy in three different stages: (a) for the first frame, we extract the key prompts related to core semantics (\eg, color, object count, relative positions) to prompt the verifiers to provide feedback that determines the consistency and correctness of subsequent frames; (b) for intermediate frames, we apply dynamic prompt in test verifiers $\mathcal{V}$ to focus on action description and motion continuity based on the context established by the first frame; (c) lastly, we prompt test verifiers to assess the overall quality of the final text-video alignment while mitigating the risk of accumulating temporal artifacts from excessive motion. To maintain smooth transitions across these distinct stages, we introduce \emph{adaptive branching} by injecting additional initial noise samples when switching between stages, thereby improving temporal coherence and diversity.

\noindent \textbf{Heuristic pruning.}
Throughout the generation process, we model the video as the dynamic growth of a forest, where trees represent possible generation paths and are expanded and pruned over time. Similar to random linear search, we start by generating $N$ initial frames, corresponding to the roots of $N$ trees. Each time step $t \in [0, T-1]$ corresponds to a layer in the tree, with each frame acting as a node. At each time step, every surviving parent node $k_{t-1}$ dynamically branches into $b_t$ candidate continuations. All $k_{t-1} \cdot b_t$ nodes are evaluated using a heuristic reward score $H$ by test verifiers $\mathcal{V}$, after which only the top $k_t$ nodes are retained for further growth. The heuristic score balances local frame quality with global consistency to prioritize the most promising paths. By iteratively applying adaptive branching and heuristic pruning, ToF search efficiently explores the search space while maintaining manageable compute costs. See Algorthm~\ref{alg:tof_search} for more details.

\noindent \textbf{Complexity analysis.}
The time complexity of growing one level of the tree is:
\begin{equation} 
O(k_{t-1} b_t+b_t\log(k_{t-1} b_t)). 
\end{equation}
Here, generating $k_{t-1} b_t$ nodes takes $O(k_{t-1} b_t)$ time, and heap sorting for pruning costs $O(b_t \log(k_{t-1} b_t))$. By iteratively applying dynamic branching and heuristic pruning, the deepest leaf nodes in the forest correspond to the final frames of the video, with the path to those nodes representing the optimal video sequence. The overall time complexity of this process is:
\begin{equation} 
\label{eq:time_complexity_origin}
O(k_0 + \sum_{t=1}^{T-1}{k_{t-1}b_t + b_t \log(k_{t-1} b_t)}). 
\end{equation}
In practice, we set $k_0 = N$ and a branching limit $b_i\leq b = 2$. In the worst-case scenario, assuming $b_i = b = 2$ for all $i$, the resulting time complexity is:

\begin{equation} 
O(N + TN + 2T \log(N)) = O(TN). \end{equation}
This complexity is consistent with the random linear search. In our practical experiments, we perform branching operations only at specific prompt stages to ensure a diverse and stable transition between stages. Consequently, $b_t$ remains 1 for most timesteps, and Eq.~\ref{eq:time_complexity_origin} can simplify to $O(N + T)$.
Compared to the quadratic complexity of random linear search, our proposed ToF significantly reduces computational costs while maintaining high sample diversity. The logarithmic dependency on $N$ ensures efficient scaling. Additionally, by dynamically adjusting the branching factor, we achieve a better trade-off between exploration in early timesteps and convergence in later stages. For detailed complexity analysis, please refer to supplementary materials.

\begin{algorithm}[!t]
\caption{Tree-of-Frames (ToF) Search}
\label{alg:tof_search}
\begin{algorithmic}[1]
\REQUIRE Initial number of roots $N$, maximum tree depth $T$, branching factors $\{b_t\}_{t=1}^T$, pruning sizes $\{k_t\}_{t=0}^T$, heuristic score $H$ by test verifier $\mathcal{V}$, video generator $\mathcal{G}$ with image-level scaling, noise distribution $\mathcal{N}$
\ENSURE Video path $\hat{v}$ with the highest heuristic score
\vspace{2mm}
\STATE Initialize empty priority queue $Q$
\FOR{$i = 1$ to $N$}
    \STATE Sample initial noise $z^{(i)} \sim \mathcal{N}$
    \STATE Inital root frame $f^{(i)}_0 \gets z^{(i)}, 0$
    \STATE Enqueue $(f^{(i)}_0, \text{score}=0, \text{path}=\{f^{(i)}_0\})$ into $Q$
\ENDFOR
\vspace{1mm}
\FOR{$t = 1$ to $T$}
    \STATE Initialize empty list $\mathcal{C} \gets \{\}$
    \FOR{$j = 0$ to $k_{t-1}$}
        \STATE Dequeue node $(f, s, p)$ from $Q$
        \FOR{$m = 1$ to $b_t$}
            \STATE Generate continuation $f_m \gets \mathcal{G}(f, t)$
            \STATE Compute heuristic reward $h_m \gets H(f_m, t)$
            \STATE Add $(f_m, s + h_m, p \cup \{f_m\})$ to $\mathcal{C}$
        \ENDFOR
    \ENDFOR
    \STATE Heap sort $\mathcal{C}$ by total score in descending order
    \STATE Clear $Q$
    \FOR{$n = 1$ to $k_t$}
        \STATE Enqueue the $n$-th top node from $\mathcal{C}$ into $Q$
    \ENDFOR
\ENDFOR
\vspace{1mm}
\STATE Final verify $(\hat{f}, \hat{s}, \hat{v}) \gets \arg\max\limits_{(f, s, v) \in \mathcal{C}} s$
\RETURN $\hat{v}$
\end{algorithmic}
\end{algorithm}

\subsection{Multi-Verifiers}
Beyond test-time scaling in policy models, previous research~\cite{mahan2024generativerewardmodels,zhang2025generativeverifiersrewardmodeling,lifshitz2025multiagentverificationscalingtesttime} has demonstrated that applying test-time scaling to generative verfier models can significantly enhance performance. This improvement can be achieved through methods such as majority voting with a single verifier model~\cite{mahan2024generativerewardmodels} or by ensembling multiple verifiers~\cite{lifshitz2025multiagentverificationscalingtesttime}.
To further boost the performance of test-time scaling in video generation, we employ a mixture of different verifiers to mitigate biases and select the best videos from the candidates: 
\begin{equation}
\begin{aligned} 
 \hat{i} &= \mathop{\arg\max}_{0 < i < n} \left( H(f^{(i)}) \right) \\ &= \mathop{\arg\max}_{0 < i < n} \left( \frac{1}{|\mathcal{M}|} \sum_{v \in \mathcal{M}}c_v \text{Rank}_v(f^{(i)})\right),
\label{eqn:aggregated-score}
\end{aligned}
\end{equation}
where $\mathcal{M}$ is the set of test verifiers, $\text{Rank}_v$ indicates the score ranking assigned by verifier $v \in \mathcal{M}$ to the $i$th candidate video $f^{(i)}$, $c_v$ denotes the weight associated with verifier $v$, $n$ is the total number of sampled candidates, and $\hat{i}$ is the index of the candidate with the highest score. This approach ensures the robustness of test-time scaling and yields better performance gains.

\section{Experiment}
\subsection{Experiment Setup}
\textbf{Video Generation Models.}
We evaluate our TTS strategy (\ie, random linear search and ToF search) using six popular open-sourced pre-trained video generation models, including three diffusion-based video models (OpenSora-v1.2~\cite{opensora}, CogVideoX-2B, and CogVideoX-5B~\cite{yang2024cogvideox}) and three autoregressive models (NOVA~\cite{deng2024NOVA}, Pyramid-Flow (SD3), and Pyramid-Flow (FLUX)~\cite{jin2024pyramid}). These models span a parameter range from 0.6B to 5B.

\noindent\textbf{Test Verifiers.}
To obtain reasonable feedback and provide the heuristic score $H$ in different stages, we leverage three multi-modal reward models specific to video generation (\ie, VisionReward~\cite{xu2024visionreward}, VideoScore~\cite{he2024videoscore}, and VideoLLaMA3~\cite{zhang2023video-llama}) to assess generated video quality under two search algorithms. VisionReward~\cite{xu2024visionreward} is designed to capture human preferences across multiple dimensions (dividing the evaluation into 29 weighted questions), while VideoScore~\cite{he2024videoscore} 
 is initialized from LMM and trained on a dataset containing human-provided multi-aspect scores to automatically assess video quality. VideoLLaMA3~\cite{zhang2023video-llama} is a multimodal foundation model that exhibits state-of-the-art image and video understanding. For comparison, we use metrics VBench~\cite{huang2024vbench} as a ground-truth "verifier" to demonstrate the upper bound achievable by the three test verifiers.

\begin{figure*}[t]
    \centering
    \includegraphics[width=\textwidth]{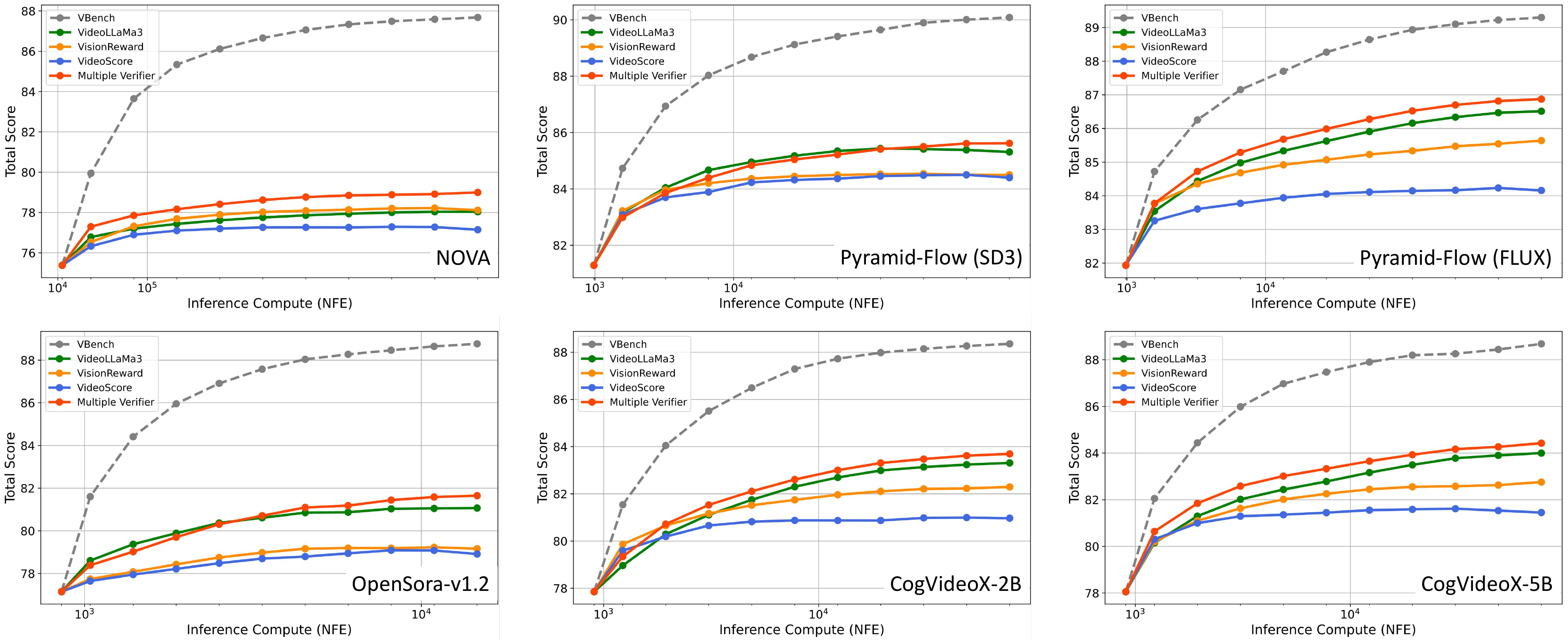}
    \caption{\textbf{Performance of random linear search on different video models and verifiers.}  The top row displays results for autoregressive models, while the bottom row shows diffusion-based models. The initial points of the curves represent the random video sample results without TTS. The models are arranged in order of increasing parameter count from left to right; different colored curves represent the performance trends under various verifiers, and the gray dashed line corresponds to the baseline established by VBench, which serves as a ground-truth verifier.}
    \label{fig:fig1}
\end{figure*}

\begin{figure*}[t]
    \centering
    \includegraphics[width=\linewidth]{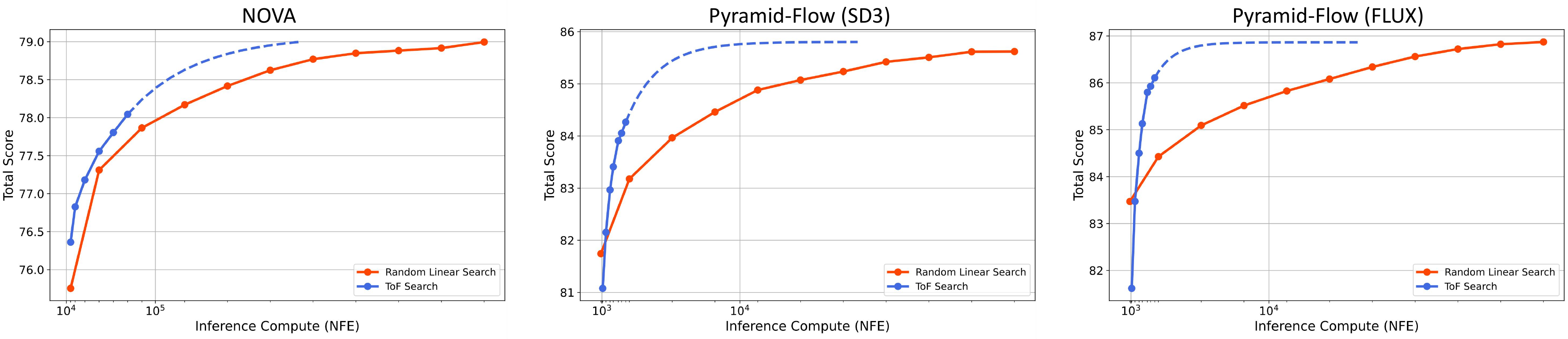}
    \caption{\textbf{Comparison between random linear search and ToF search.} The red curve represents random linear search. The blue curve represents ToF search, with the dashed line being the predicted curve from a geometric series decay approximation. Curve fitting reveals that similar subsequent trends tend to converge to an upper limit.}
    \label{fig:fig2}
\end{figure*}

\noindent\textbf{Details of Search Algorithms.}
We conduct experiments on two search algorithms assessed on VBench~\cite{huang2024vbench}. For the random linear search, experiments are conducted on 6 video generation models using various verifiers, where the initial sample noise level is incremented from 1 to 30 for each trial. In the case of ToF search, the method is applied to 3 autoregressive models using the best-performing multiple verifier where the initial sample noise is varied from 1 to 7.

\noindent\textbf{Metrics.}
To quantify the performance of text-to-video generation, we use VBench~\cite{huang2024vbench}, which is a comprehensive benchmark incorporating 16 fine-grained dimensions that evaluate both motion quality and semantic alignment. 
For the computational cost, we extend the metric of the number of function evaluations (NFE)~\cite{ma2025inferencetimescalingdiffusionmodels,tian2024visual,yu2023language} from image generation to video generation by defining NFE as the product of the total number of denoising steps executed during the generation process and the temporal length of latent embeddings, which takes the temporal dimension of the video into inference computational costs.

\subsection{Analysis of Experimental Results}
{\textbf{TTS consistently yields stable performance gains across different video generation models.}}
We conduct a series of random linear search experiments across multiple video generation models using different verifiers. In these experiments, the final video outputs were evaluated with the VBench~\cite{huang2024vbench} total score—a composite metric aggregating 16 distinct dimensions of quality (\eg, motion smoothness, semantic alignment, aesthetic quality). Figure~\ref{fig:fig1} demonstrates that as the inference computational budget increases, all video generation models exhibit improved performance across different verifiers, eventually approaching a convergence limit once a certain threshold is reached. This finding indicates that the TTS strategy can effectively guide the search process during test time and significantly enhance generation quality. Moreover, when comparing different verifiers applied to the same video model, we observe varying growth rates and extents in their performance curves. 
This divergence suggests that each verifier emphasizes different evaluation aspects. 

\noindent \textbf{Multiple verifiers can further boost the curve of TTS.} Beyond the test-time scaling in video generation models, we ensemble the multiple verifiers in Figure~\ref{fig:fig1} that can further boost the performance of test-time scaling in video generation. Such a mixture of different verifiers can also mitigate biases and select the best video from the candidates.

\noindent {\textbf{Advanced foundation models offer significant potential for improvement with TTS.}} Additionally, comparative analysis across video models in Figure~\ref{fig:fig1} and Table~\ref{tab: bigtable} reveals that lightweight models (\eg, NOVA) exhibit only marginal performance improvements with increased inference effort, whereas larger models (\eg, CogVideoX-5B) benefit from a substantially wider search space and thus achieve more significant enhancements. This observation underscores the potential of larger models to leverage the TTS strategy more effectively, thereby yielding higher-quality video generation under increased computational budgets.

\begin{table}[h]
  \caption{Inference-time scaling cost comparison on GFLOPs.}
  \vspace{-5pt}
  \footnotesize
  \centering
    \begin{tabular}{l|>{\centering\arraybackslash}m{2cm}|>{\centering\arraybackslash}m{2cm}}
      \toprule
      \textbf{Methods} & \textbf{Linear Search} & \textbf{ToF Search} \\
      \midrule
      \textbf{Pyramid-Flow(FLUX)}& $5.22\times 10^7$ & $1.62 \times 10^7$ \\  
      \textbf{Pyramid-Flow(SD3)} & $3.66 \times 10^7$  & $1.13 \times 10^7$ \\
      \textbf{NOVA}  & $4.02\times 10^6$ & $1.41\times10^6$ \\
      \bottomrule
    \end{tabular}
    \label{tab:flops}
\end{table}

\begin{figure*}[t]
    \centering
    \includegraphics[width=\linewidth]{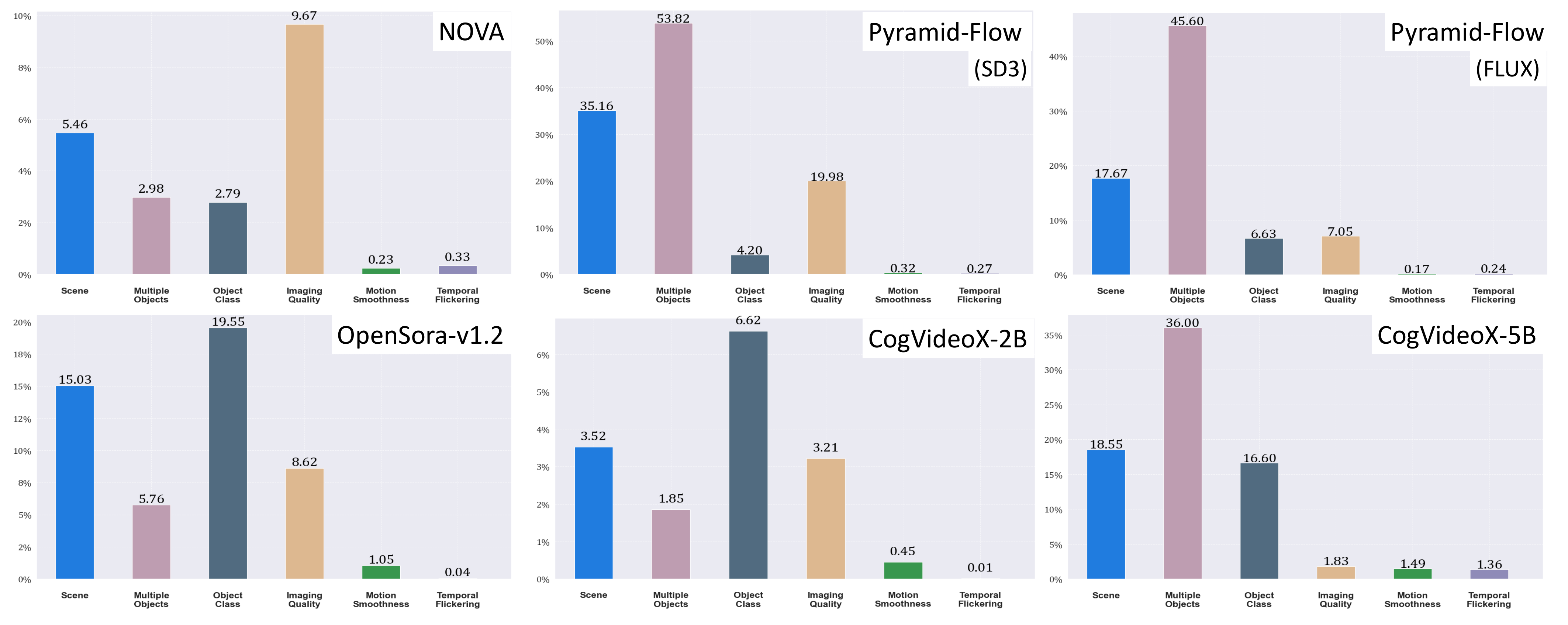}
    \caption{\textbf{Qualitative TTS performance improvement ratio on different complexities of prompts} across different video generation models across diverse benchmark dimensions of Vbench.}
    \label{fig:fig3}
\end{figure*}

\definecolor{+}{rgb}{0.70,0.13,0.13}
\definecolor{-}{rgb}{0.25,0.41,0.88}
\begin{table*}[t]
  \caption{\textbf{Quantitative Performance Comparison} on VBench across different video generation models.}
  \label{tab: bigtable}
  \vspace{-10pt}
  \small
  \resizebox{\textwidth}{!}{
    \begin{tabular}{l|>{\centering\arraybackslash}m{2.4cm}|>{\centering\arraybackslash}m{2.4cm}|>{\centering\arraybackslash}m{2.4cm}|>{\centering\arraybackslash}m{2.4cm}|>{\centering\arraybackslash}m{2.4cm}|>{\centering\arraybackslash}m{2.4cm}}
      \toprule
      \textbf{}          & \textbf{ Total Score}  & \textbf{ Quality Score} & \textbf{ Semantic Score} & \textbf{ Object Class} & \textbf{ Scene } & \textbf{ Multiple Objects } \\
      \midrule
      \textit{Diffusion-based Models} & & & & & & \\
      \textbf{CogVideoX-5B}   & 81.61  &  82.75 & 77.04  & {85.23} & { 53.20} & {62.11}  \\
      \textbf{\hspace{2cm}+ TTS} &
      \hspace{0.75cm}{84.42}\textsuperscript{{\scriptsize\color{+}{+3.44\%}}}   & 
      \hspace{0.75cm}{84.32}\textsuperscript{{\scriptsize\color{+}{+1.90\%}}}  & 
      \hspace{0.75cm}\textbf{84.83}\textsuperscript{{\scriptsize\color{+}{+10.1\%}}}   & 
      \hspace{0.75cm}{99.38}\textsuperscript{{\scriptsize\color{+}{+16.6\%}}} & 
      \hspace{0.75cm}\textbf{63.07}\textsuperscript{{\scriptsize\color{+}{+18.6\%}}} & 
      \hspace{0.75cm}{84.47}\textsuperscript{{\scriptsize\color{+}{+36.0\%}}} \\
      \midrule
      \textbf{CogVideoX-2B}   & 80.91  & 
      82.18 
      & 75.83  & 83.37    &  51.14   &   62.63    \\
      \textbf{\hspace{2cm}+ TTS} & 
      \hspace{0.75cm}{83.89}\textsuperscript{{\scriptsize\color{+}{+3.68\%}}} & 
      \hspace{0.75cm}{85.27}\textsuperscript{{\scriptsize\color{+}{+3.76\%}}}   & 
      \hspace{0.75cm}{78.39}\textsuperscript{{\scriptsize\color{+}{+3.38\%}}}  & 
      \hspace{0.75cm}{88.89}\textsuperscript{{\scriptsize\color{+}{+6.62\%}}}   & 
      \hspace{0.75cm}{52.94}\textsuperscript{{\scriptsize\color{+}{+3.52\%}}}   & 
      \hspace{0.75cm}{63.79}\textsuperscript{{\scriptsize\color{+}{+1.85\%}}}   \\
      \midrule
      \textbf{OpenSora-v1.2}     & 79.76  & 81.35 & 73.39  & {82.22} & {42.44} & {63.34}  \\
      \textbf{\hspace{2cm}+ TTS} & \hspace{0.75cm}{81.65}\textsuperscript{{\scriptsize\color{+}{+2.37\%}}}    & \hspace{0.75cm}{81.90}\textsuperscript{{\scriptsize\color{+}{+0.68\%}}}  & \hspace{0.75cm}{80.63}\textsuperscript{{\scriptsize\color{+}{+9.87\%}}}   &
      \hspace{0.75cm}{98.29}\textsuperscript{{\scriptsize\color{+}{+19.5\%}}}  & 
      \hspace{0.75cm}{48.82}\textsuperscript{{\scriptsize\color{+}{+15.0\%}}}  & 
      \hspace{0.75cm}{66.99}\textsuperscript{{\scriptsize\color{+}{+5.76\%}}}  \\
      
      \midrule
      \midrule
      \textit{Autoregressive Models} & & & & & & \\
      \textbf{Pyramid-Flow (SD3)}  & 81.72 & 84.74 & 69.62  & {86.67} & {43.20} & {50.71}    \\
      \textbf{\hspace{2cm}+ TTS} & 
      \hspace{0.75cm} {85.31}\textsuperscript{{\scriptsize\color{+}{+4.39\%}}} 
      & 
      \hspace{0.75cm} {86.84}\textsuperscript{\scriptsize\color{+}{+2.48\%}}  
      & \hspace{0.75cm}{79.21}\textsuperscript{\scriptsize\color{+}{+13.8\%}}  
      & \hspace{0.75cm}{90.31}\textsuperscript{{\scriptsize\color{+}{+4.20\%}}}
      & \hspace{0.75cm}{58.39}\textsuperscript{{\scriptsize\color{+}{+35.2\%}}}
      & \hspace{0.75cm}{78.00}\textsuperscript{{\scriptsize\color{+}{+53.8\%}}}\\
      \midrule
      \textbf{Pyramid-Flow (FLUX)}  & 81.61 & 84.11 & 71.61 & {93.49} & {47.65} & {61.08}    \\
      \textbf{\hspace{2cm}+ TTS} & 
      \hspace{0.75cm}\textbf{86.51}\textsuperscript{\scriptsize\color{+}{+5.86\%}}
      & \hspace{0.75cm}\textbf{87.50}\textsuperscript{\scriptsize\color{+}{+3.26\%}}
      & \hspace{0.75cm}{82.56}\textsuperscript{\scriptsize\color{+}{+18.6\%}}
      & \hspace{0.75cm}\textbf{99.69}\textsuperscript{{\scriptsize\color{+}{+6.63\%}}}
      & \hspace{0.75cm}{56.07}\textsuperscript{{\scriptsize\color{+}{+17.7\%}}}
      & \hspace{0.75cm}\textbf{88.93}\textsuperscript{{\scriptsize\color{+}{+45.6\%}}} \\
      \midrule
      \textbf{NOVA}        & 78.56  & 83.79 & 57.63  & {91.36} & {45.22} & {67.87}  \\
      \textbf{\hspace{2cm}+ TTS} & \hspace{0.75cm}{79.80}\textsuperscript{\scriptsize\color{+}{+1.58\%}}   & \hspace{0.75cm}{84.99}\textsuperscript{\scriptsize\color{+}{+1.43\%}}    & \hspace{0.75cm}{59.03}\textsuperscript{\scriptsize\color{+}{+2.43\%}}   & 
      \hspace{0.75cm}{93.91}\textsuperscript{{\scriptsize\color{+}{+2.79\%}}} & 
      \hspace{0.75cm}{47.69}\textsuperscript{{\scriptsize\color{+}{+5.46\%}}} & 
      \hspace{0.75cm}{69.89}\textsuperscript{{\scriptsize\color{+}{+2.98\%}}} \\
      \bottomrule
    \end{tabular}}
  \vspace{-5pt}
\end{table*}

\definecolor{+}{rgb}{0.70,0.13,0.13}
\definecolor{-}{rgb}{0.25,0.41,0.88}

\noindent {\textbf{ToF Search is more efficient and superior to the random linear search.}} We implement the ToF search in three autoregressive models and conduct a comparison experiment with the random linear search and ToF search in Figure~\ref{fig:fig2}.  We observe that the ToF search achieves comparable performance at a much lower computational cost, highlighting its high efficiency. To minimize the significant differences in computational costs among models of different sizes, We also show quantitative results of GFLOPs in Table~\ref{tab:flops}.  Moreover, larger and better models show higher efficiency, as evidenced by the faster rising speed of the curve.

\noindent {\textbf{Performance across most dimensions can be greatly improved with TTS.}}
The complexity of prompts across diverse benchmark dimensions is a key component in video TTS. We conduct experiments to quantitatively evaluate the performance improvement of different models using TTS methods across various dimensions (See Figure~\ref{fig:fig3} and Table~\ref{tab: bigtable}). As ToF and random linear search can achieve a similar convergence score during test-time scaling, we choose the better score for (+TTS).
We find that for common prompt sets (\eg, Scene, Object) and easily assessable categories (\eg, Imaging Quality), TTS methods achieve significant improvements across different models. 

\noindent {\textbf{A few dimensions heavily rely on the capabilities of foundation models, making improvements challenging for TTS.}}
 However, for some hard-to-evaluate latent properties (\eg, Motion Smoothness, Temporal Flickering), the improvement is less pronounced. This is likely because Motion Smoothness requires precise control of motion trajectories across frames, which is challenging for current video generation models to achieve. Temporal Flickering, on the other hand, involves maintaining consistent appearance and intensity over time, which is difficult to precisely assess, especially when dealing with complex scenes and dynamic objects. (See Figure~\ref{fig:fig3} and Table~\ref{tab: bigtable})
\section{Conclusion}

In conclusion, this study presents a novel framework for test-time scaling in video generation, redefining it as a search problem for optimal video trajectories. We build the search space in TTS by test-time verifiers and to provide feedback and employ heuristic algorithms like random linear search and the more efficient ToF search algorithm. Extensive experiments demonstrate that scaling the search space can boost the video performance across various video generation models, and our proposed ToF search can significantly reduce scaling cost when achieving high-quality video outputs. This framework opens new avenues for research into efficient test-time optimization strategies in video generation.

\noindent \textbf{Acknowledgments:} This work was supported in part by
the National Natural Science Foundation of China under
Grant 62206147, and in part by 2024 Tencent AI Lab
Rhino-Bird Focused Research Program.


{
    \small
    \bibliographystyle{ieeenat_fullname}
    \bibliography{main}
}
\appendix
\newpage
\section{More Implementation Details}

\subsection{More Discussion of Image-level Alignment}
In our generation process, inspired by recent work~\cite{guo2025image-cot}, each frame is generated with image-level TTS. Specifically, we employ a two-stage evaluation mechanism. First, a potential assessment reward model examines whether the partially generated frame exhibits sufficient visual clarity for meaningful evaluation and assigns a binary label. If the output is deemed unclear (`no'), the model skips further processing for that frame. Otherwise (`yes'), the frame advances to a secondary evaluation stage where its potential to yield a high-quality final image is assessed. Again, a binary decision is made: if the frame is unlikely to lead to an optimal outcome, the generation path is truncated immediately; if it passes, the synthesis continues to produce the final image.

\subsection{More Discussion of Hierarchical Prompting}
In our approach to hierarchical prompting, we employ DeepSeek-R1-8B~\cite{guo2025deepseek}, a large language model distilled from the LLaMA-8B~\cite{touvron2023llama} architecture. DeepSeek-R1-8B is used to decompose a given input prompt into three distinct hierarchical prompts, each tailored to represent a specific stage of the video sequence:
\begin{itemize}
\item Static scene description: The first prompt provides a detailed depiction of the static scene intended for the initial frame, establishing the starting visual context of the sequence.
\item Action/motion directions: The second prompt outlines the actions or motion directions that guide the dynamic progression across the intermediate frames, ensuring a coherent evolution of the scene.
\item Expected ending state: The third prompt delineates the expected ending state for the concluding frame, defining the desired outcome of the video sequence.
\end{itemize}
DeepSeek-R1-8B processes the input prompt and generates these three hierarchical prompts, which are then returned as an ordered list of length three. At each stage of the video sequence—initial, intermediate, and final—the test verifier evaluates the generated output with the corresponding hierarchical prompt. This stage-specific assessment ensures that the video content aligns with the intended descriptions, maintaining both accuracy and coherence throughout the sequence. Figure~\ref{fig:reasoning_grounding} gives an example of hierarchical prompting generation and test-time verification.

\begin{figure*}[t]
    \centering
    \includegraphics[width=\linewidth]{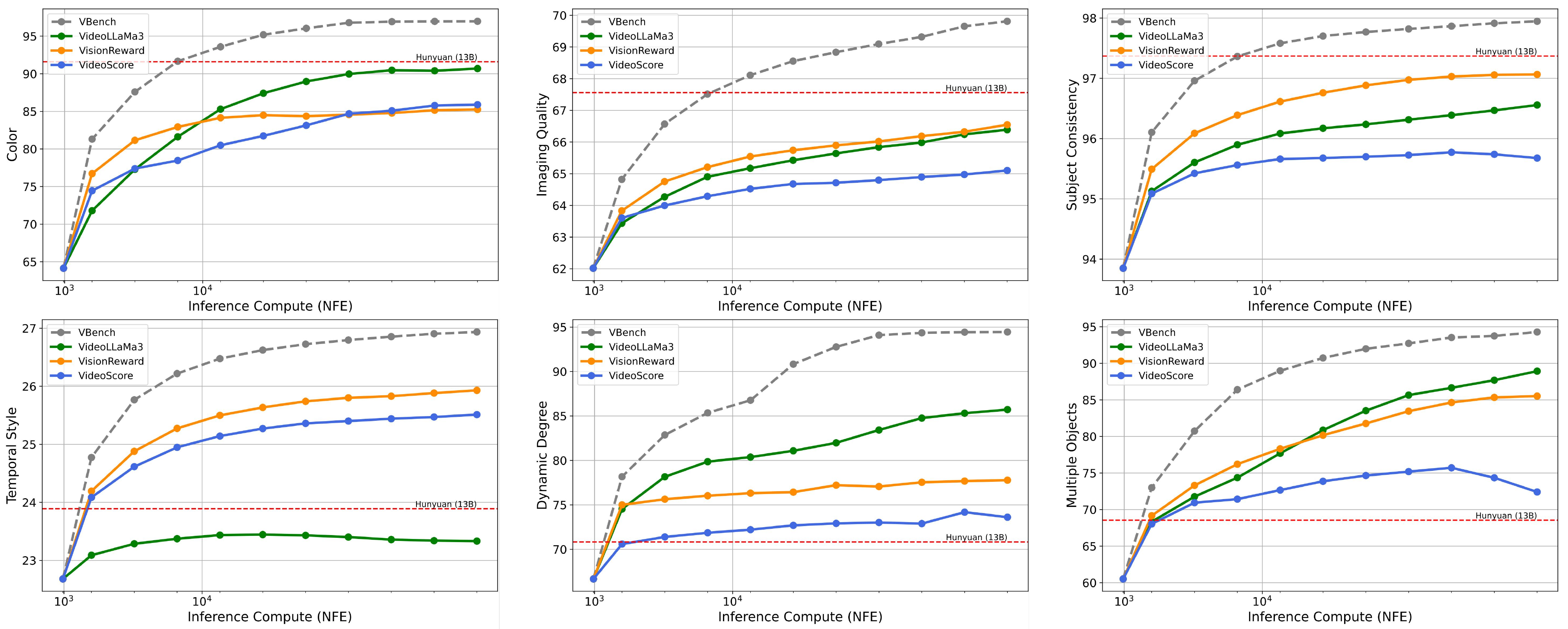}
    \caption{Using TTS, the small model (Pyramid-Flow) achieves scores that are close to, or even exceed, those of the 13B large model (HunyuanVideo) in many dimensions. The gray dashed horizontal line in the figures indicates HunyuanVideo's score in that dimension.}
    \label{fig:hunyuan_comparison}
\end{figure*}

\subsection{Detailed Complexity Analysis}
\label{supp:complexity}
In our method, the video generation process is modeled as the dynamic growth of a forest, where the trees are branched and pruned over time. Specifically, similar to the linear search, we begin by generating $N$ initial frames, representing the roots of $N$ trees in the forest. Each time step $t \in [0, T-1]$ corresponds to a level in the tree, and each frame is treated as a tree node. We consider the process in which each of the $N$ trees grows by adding one level of nodes. At each time step, the $k_{t-1}$ surviving parent nodes dynamically branch into $b_t$ possible continuations. We then evaluate all $k_{t-1} \cdot b_t$ nodes using a heuristic reward function $H$, followed by pruning to retain only the top $k_t$ branches. The time complexity of growing one level of the tree is:
\begin{equation} 
O(k_{t-1} b_t+b_t\log(k_{t-1} b_t)). 
\end{equation}
Here, generating $k_{t-1} b_t$ nodes takes $O(k_{t-1} b_t)$ time, the evaluation cost per node is $O(1)$, and the total evaluation time is $O(k_{t-1} b_t)$. Heap sorting for pruning costs $O(b_t \log(k_{t-1} b_t))$. By iteratively applying dynamic branching and heuristic pruning, the deepest leaf nodes in the forest correspond to the final frames of the video, with the path to those nodes representing the optimal video sequence. The overall time complexity of this process is:
\begin{equation} 
\label{eq:supp_time_complexity_origin}
O(k_0 + \sum_{t=1}^{T-1}{k_{t-1}b_t + b_t \log(k_{t-1} b_t)}). 
\end{equation}

In practice, we set a branching limit $b$ for dynamic branching, \ie, $b_t \leq b$. In the heuristic pruning step, we use the heuristic reward function $H$ to prune branches that fall below the average value. On average, each pruning step retains $k_t \approx \frac{k_{t-1} b_t}{2}=\frac{k_0 \prod_{i=1}^tb_i}{2^t}\leq \frac{k_0 b^t}{2^t}$ branches before $k_t$ drops to 1. Therefore, we have:
\begin{equation}
\begin{aligned} 
\label{eq:supp_time_complexity}
&\quad O(k_0 + \sum_{t=1}^{T-1}{k_{t-1} b_t + b_t \log(k_{t-1} b_t)}) \\ 
&=O(k_0 + \sum_{t=1}^{T-1}{\frac{k_0 \prod_{i=1}^tb_i}{2^t} +b_t\log(\frac{k_0 \prod_{i=1}^tb_i}{2^t})}
\end{aligned}
\end{equation}

In practice, we set $k_0 = N$ and $b = 2$. In the worst-case scenario, assuming $b_i = b = 2$ for all $i$, the resulting time complexity is:

\begin{equation} 
O(N + TN + 2T \log(N)) = O(TN). \end{equation}

This complexity is consistent with that of the linear search. However, in our actual experiments, we perform branching operations only at specific prompt stages to ensure a diverse and stable transition between stages. Consequently, $b_t$ remains 1 for most timesteps, leading to the following update rule:

\begin{equation} 
k_t= 
\left\{
\begin{array}{lc} \frac{k_{t-1} b_t}{2}, & 0 < t \leq \log(k_0) \\
1, & t > \log(k_0). \end{array} 
\right. 
\end{equation}

Thus, Eq.~\ref{eq:supp_time_complexity_origin} can simplify to:

\begin{small} 
\begin{equation}
\label{eq:supp_average_time_complexity}
\begin{aligned} 
&
\quad O(k_0 + \sum_{t=1}^{T-1}{k_{t-1} b_t + b_t \log(k_{t-1} b_t)}) \\ 
&=O(k_0 + \sum_{t=\log(k_0)+1}^{T-1}b_t + \sum_{t=1}^{\log(k_0)}{\frac{k_0 \prod_{i=1}^tb_i}{2^t} + b_t \log(\frac{k_0 \prod_{i=1}^tb_i}{2^t})}) \\ 
&=O(k_0 + T-\log(k_0) + k_0 + \log^2(k_0)-\frac{\log2}{2}log^2(k_0)) \\ 
&=O(N + T). 
\end{aligned}
\end{equation}
\end{small}

Compared to the quadratic complexity of linear search, our approach converge at a geometric rate, ultimately achieving linear complexity. It significantly reduces computational costs while maintaining high sample diversity. The logarithmic dependency on $N$ ensures efficient scaling, making our method more suitable for high-dimensional video generation. Additionally, by dynamically adjusting the branching factor, we achieve a better trade-off between exploration in early timesteps and convergence in later stages.

\section{More Experiments}
\subsection{More Quantitative Results}
We performed multiple random linear search experiments across various video generation models with different verifiers. Figures~\ref{fig:app1}-\ref{fig:app3} present more quantitative results on VBench across different dimensions. These indicate that TTS consistently delivers stable performance improvements across dimensions. Moreover, the evaluation accuracy of different verifiers varies across dimensions, justifying our use of multiple verifiers.

\subsection{More Qualitative Results}
We provide additional visual results for Pyramid-Flow~\cite{jin2024pyramid} and CogVideoX-5B~\cite{yang2024cogvideox} in Figures~\ref{fig:visual_comparison}-\ref{fig:visual_comparison_5}. Each example displays different video outputs as NFE increases with the same prompt input. The results show that video quality and text alignment improve with more TTS samples.

\subsection{Comparison with Large Models}
Figure~\ref{fig:hunyuan_comparison} compares the outputs of the small model Pyramid-Flow (2B) with TTS and the large model HunyuanVideo (13B) single. As scaling increases, the small model's performance approaches or even surpasses that of the large model in many dimensions.

\subsection{Failure Cases}
\begin{figure}[h]
    \centering
    \includegraphics[width=\linewidth]{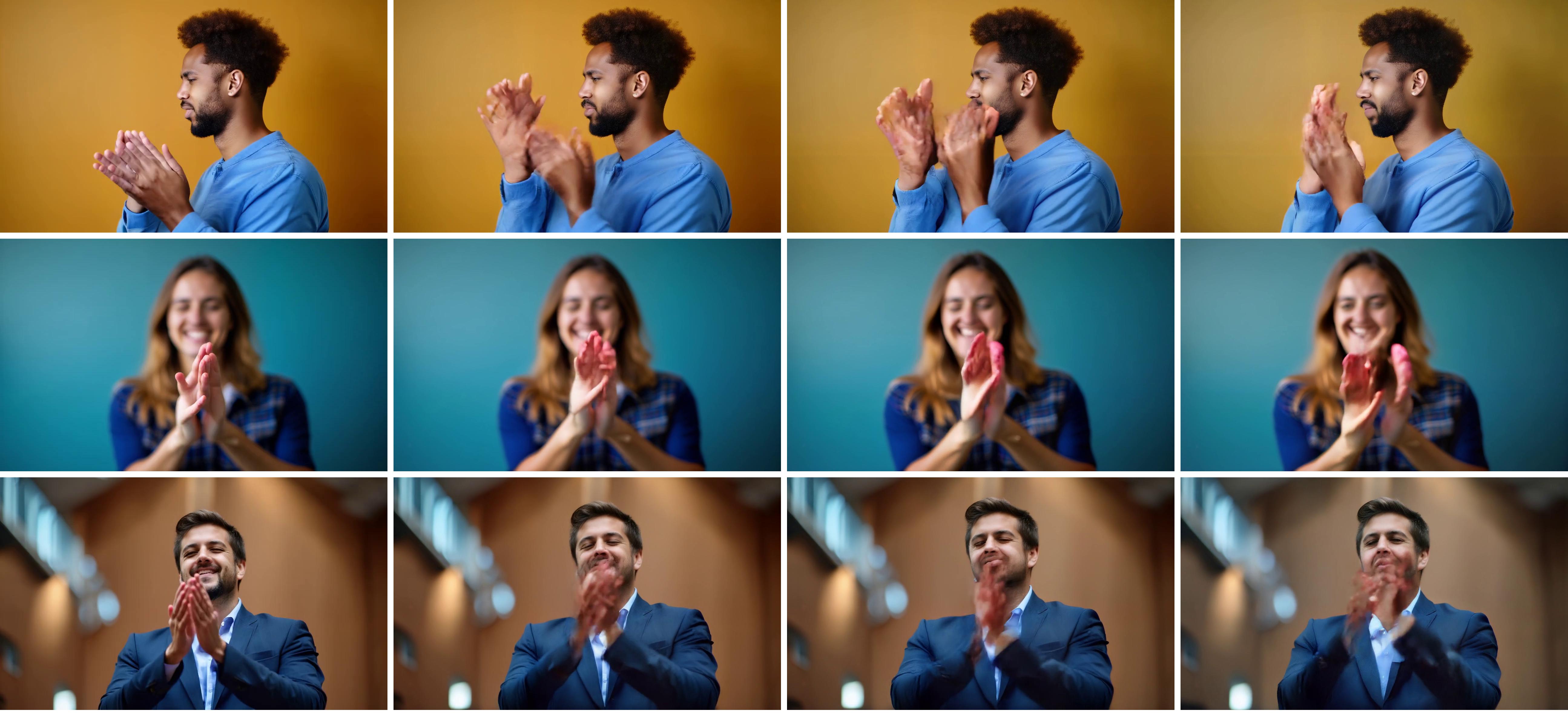}
    \caption{Failure cases on prompt ``A person is clapping''.}
    \label{fig:failure_cases}
\end{figure}

\begin{figure*}[t]
    \centering
    \includegraphics[width=0.8\linewidth]{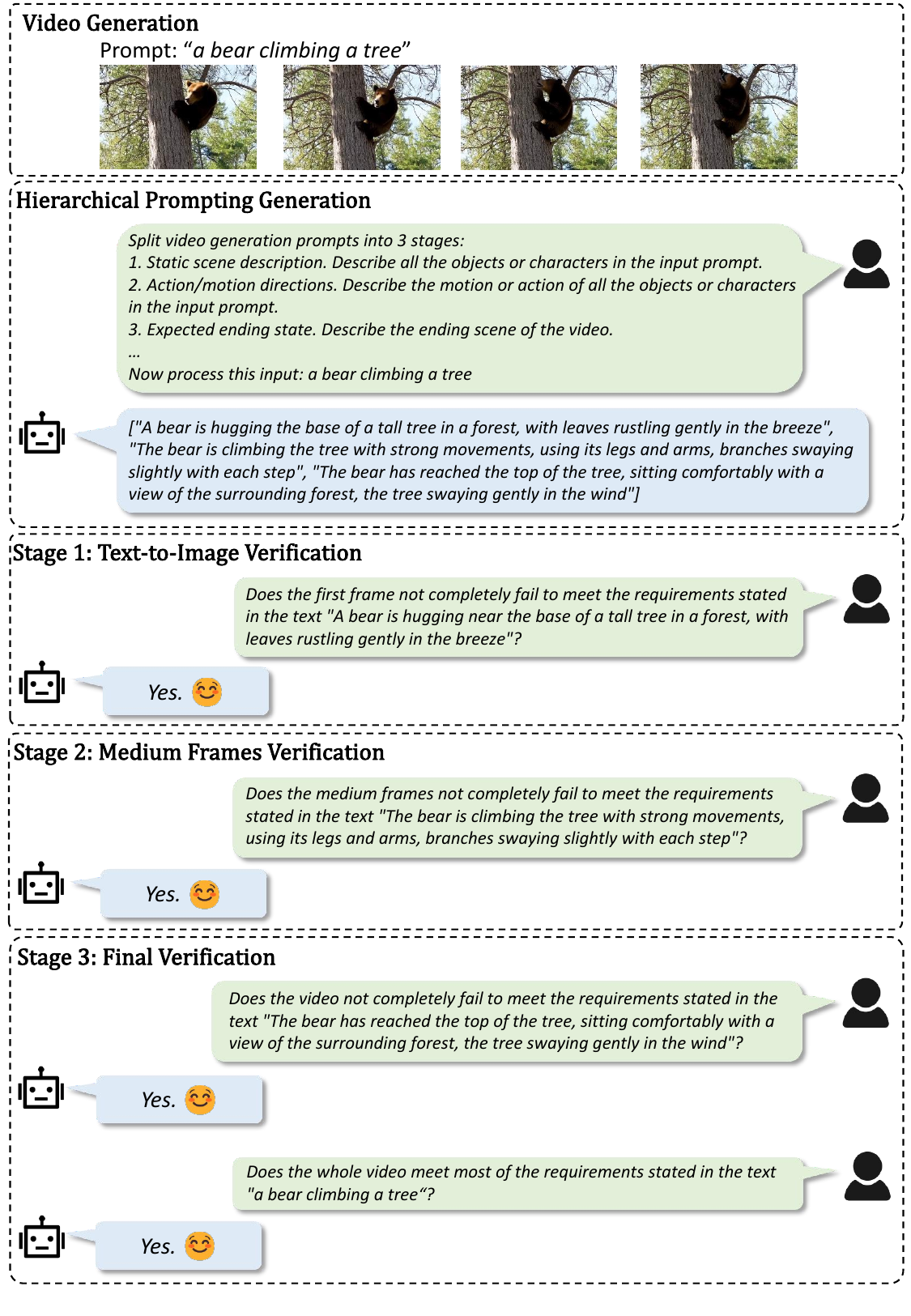}
    \caption{Verifications during TTS process.}
    \label{fig:reasoning_grounding}
\end{figure*}

\begin{figure*}[t]
    \centering
    \includegraphics[width=0.9\linewidth]{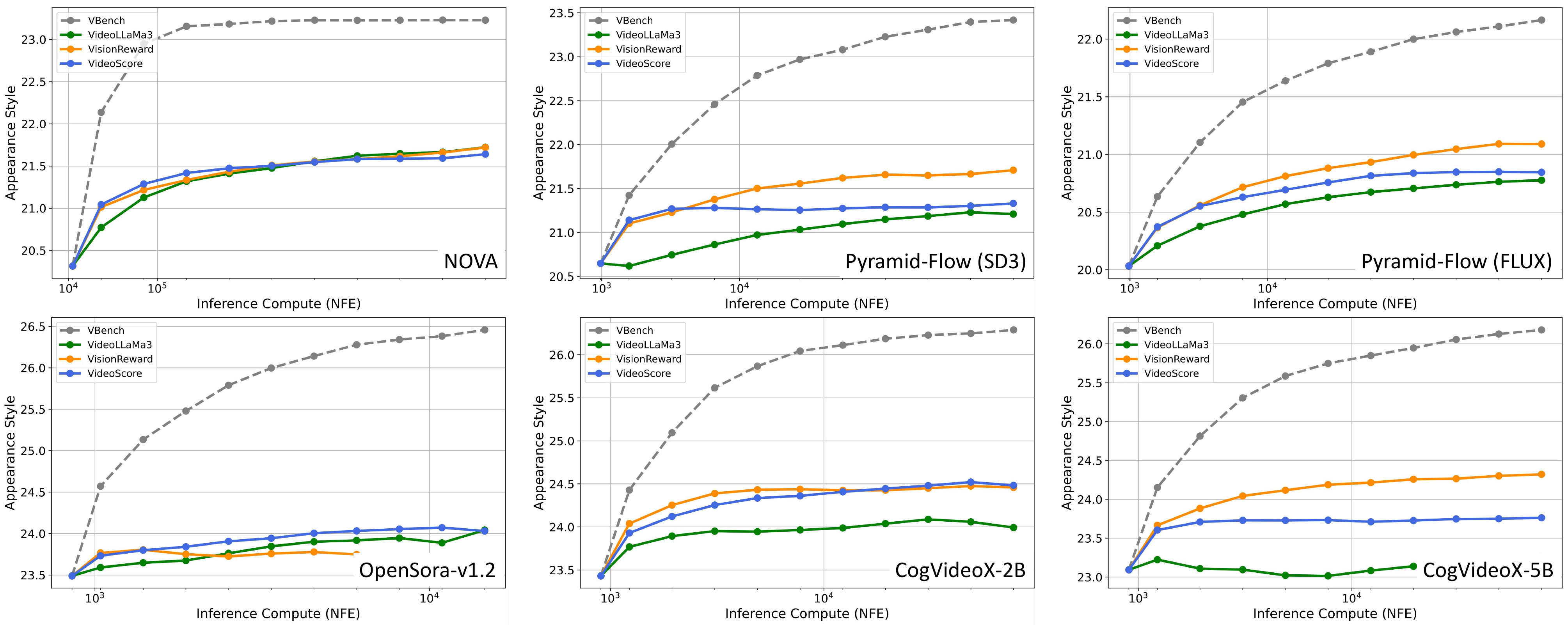}
    \vspace{-1em}
    \caption{TTS performance on Appearance Style across diverse verifiers.}
    \label{fig:app1}
\end{figure*}

\begin{figure*}[!h]
    \centering
    \includegraphics[width=0.9\linewidth]{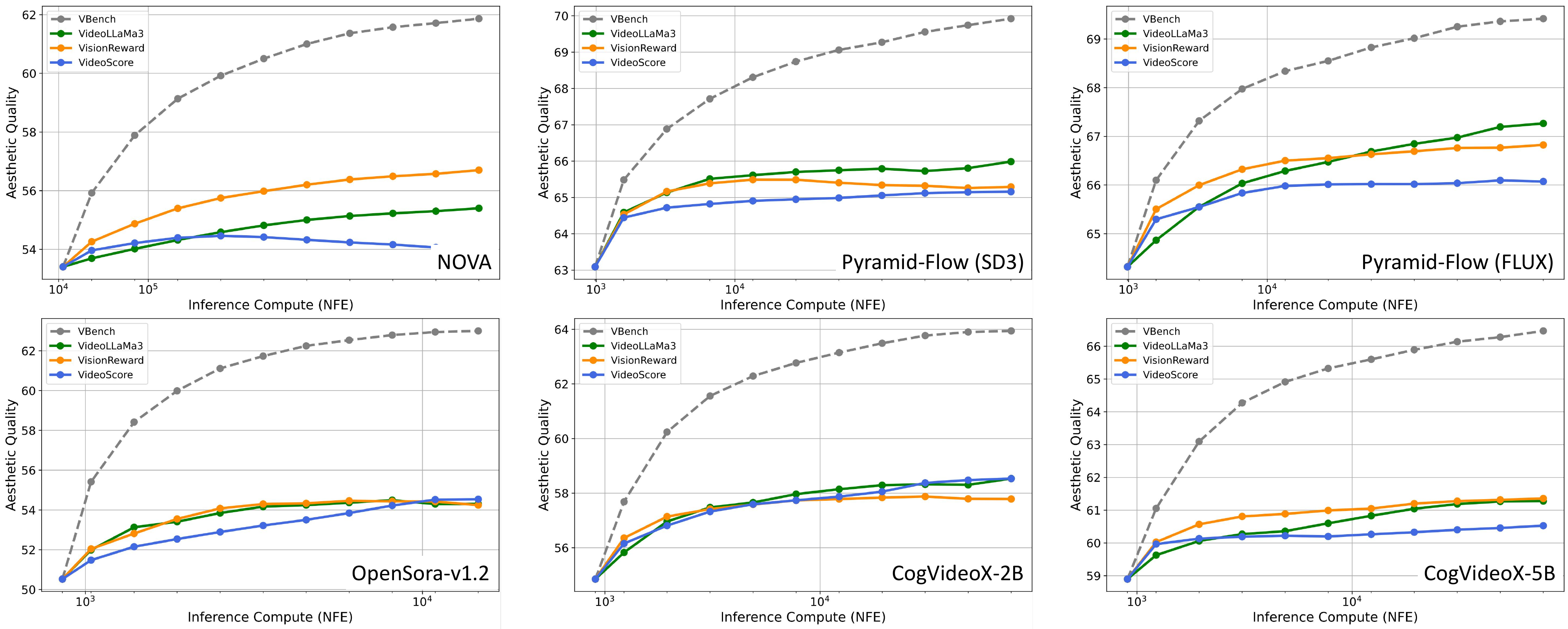}
    \vspace{-1em}
    \caption{TTS performance on Aesthetic Quality across diverse verifiers.}
    \label{fig:app2}
\end{figure*}

\begin{figure*}[!h]
    \centering
    \includegraphics[width=0.9\linewidth]{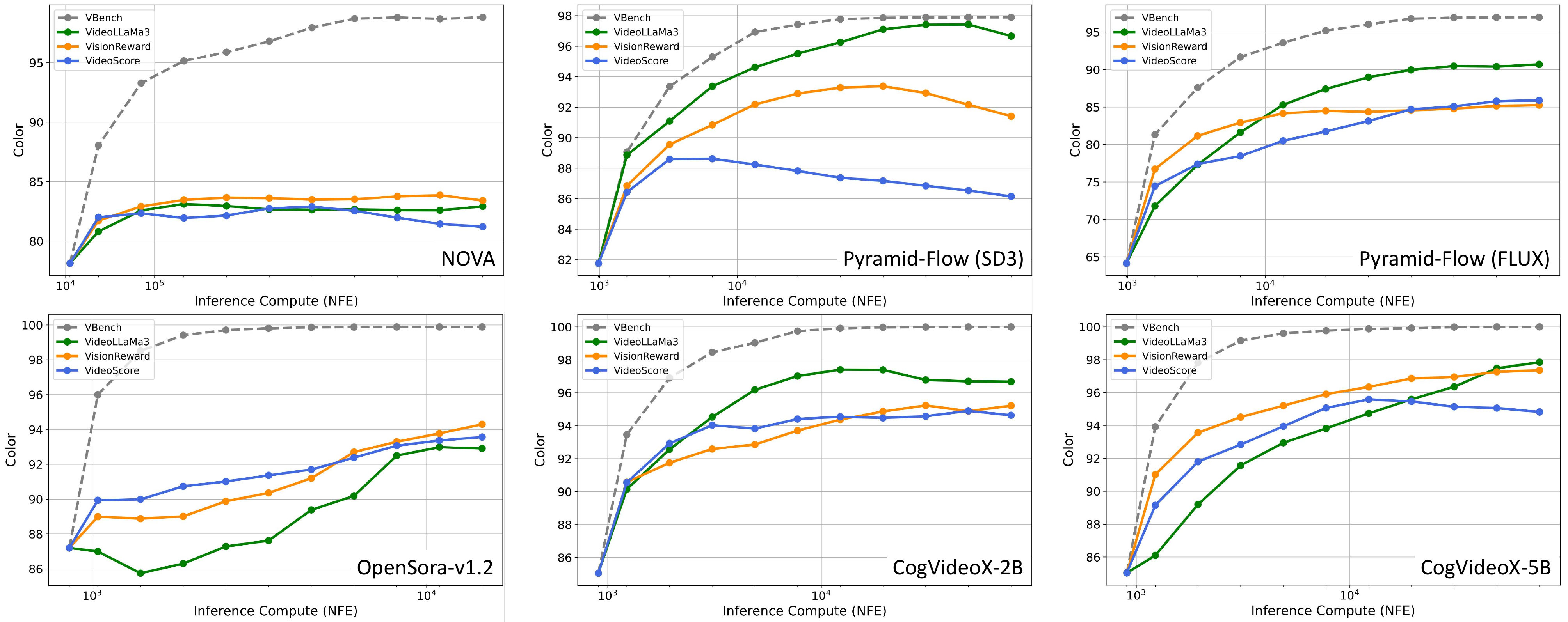}
    \vspace{-1em}
    \caption{TTS performance on Color across diverse verifiers.}
    \label{fig:app3}
\end{figure*}

\begin{figure*}[t]
    \centering
    \includegraphics[width=0.9\linewidth]{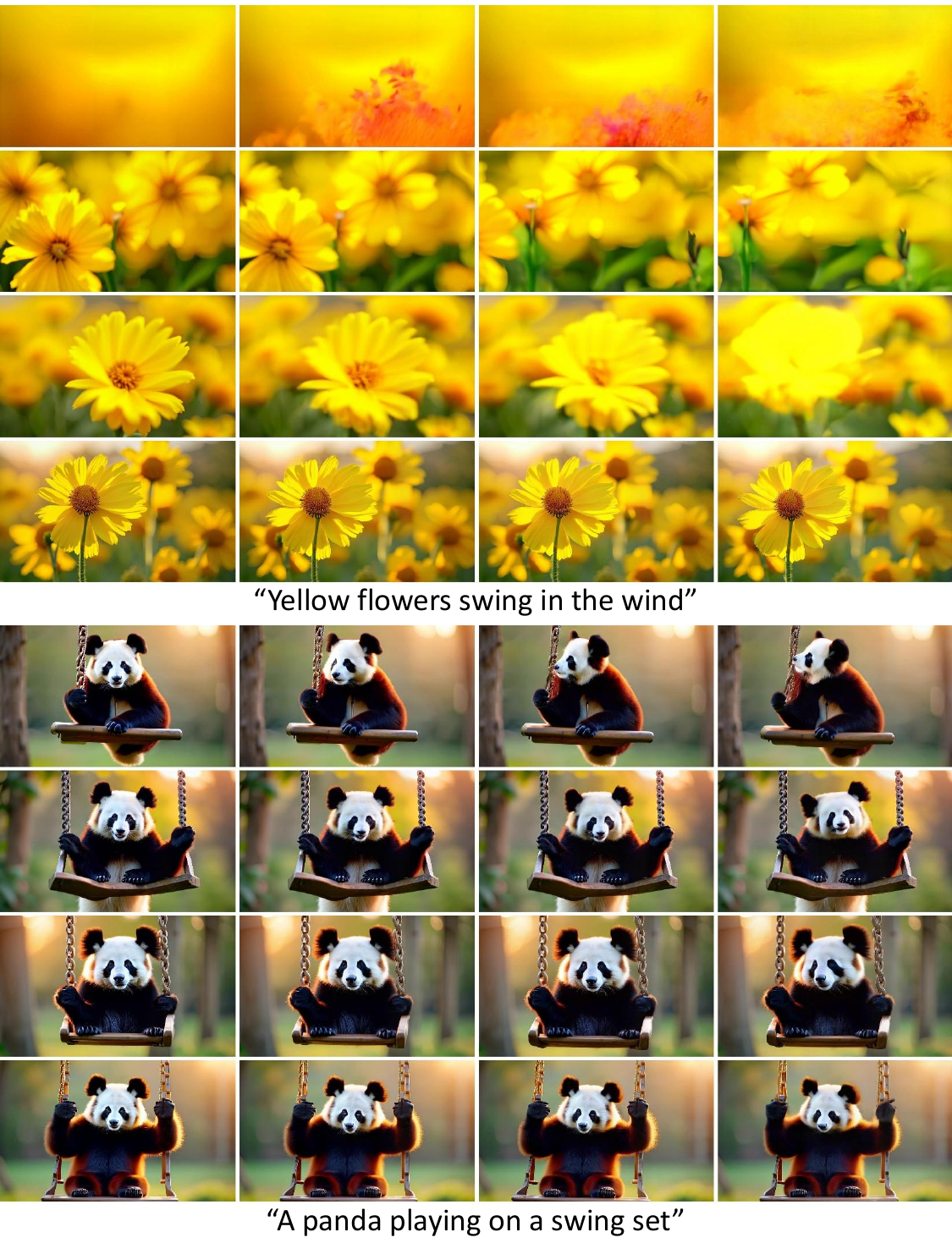}
    \caption{More visual results during TTS process on Pyramid-Flow. From left to right, each row of frames are extracted from a video sequence. From top to bottom, each row represents the output video results of TTS with an increasing number of samples.}
    \label{fig:visual_comparison}
\end{figure*}

\begin{figure*}[t]
    \centering
    \includegraphics[width=0.9\linewidth]{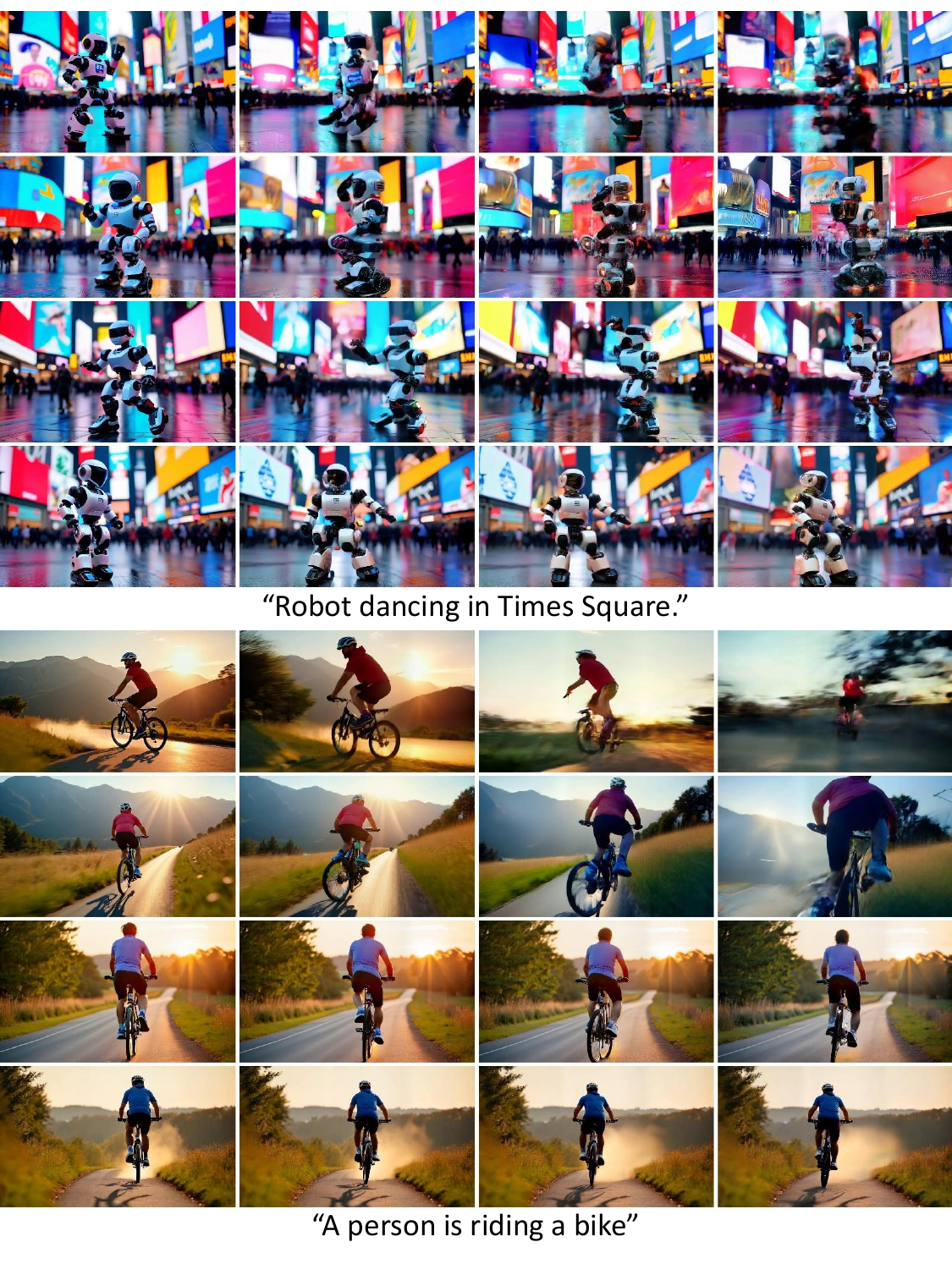}
    \caption{More visual results during TTS process on Pyramid-Flow. The TTS method can effectively alleviate common issues in video generation, such as those related to human motion and complex movements.}
    \label{fig:visual_comparison_1}
\end{figure*}

\begin{figure*}[t]
    \centering
    \includegraphics[width=0.9\linewidth]{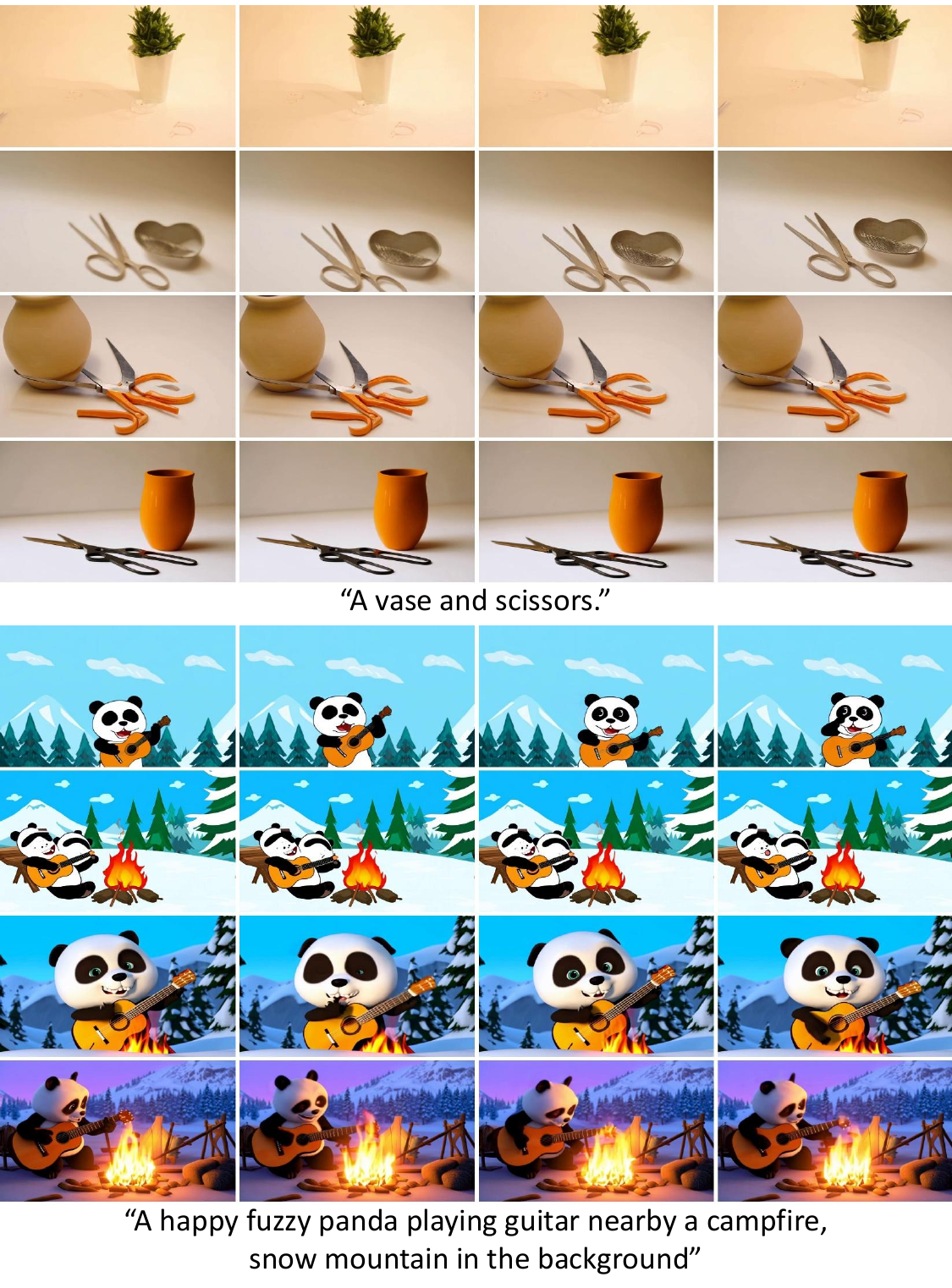}
    \caption{More visual results during TTS process on CogVideoX-5B.}
    \label{fig:visual_comparison_2}
\end{figure*}

\begin{figure*}[t]
    \centering
    \includegraphics[width=0.9\linewidth]{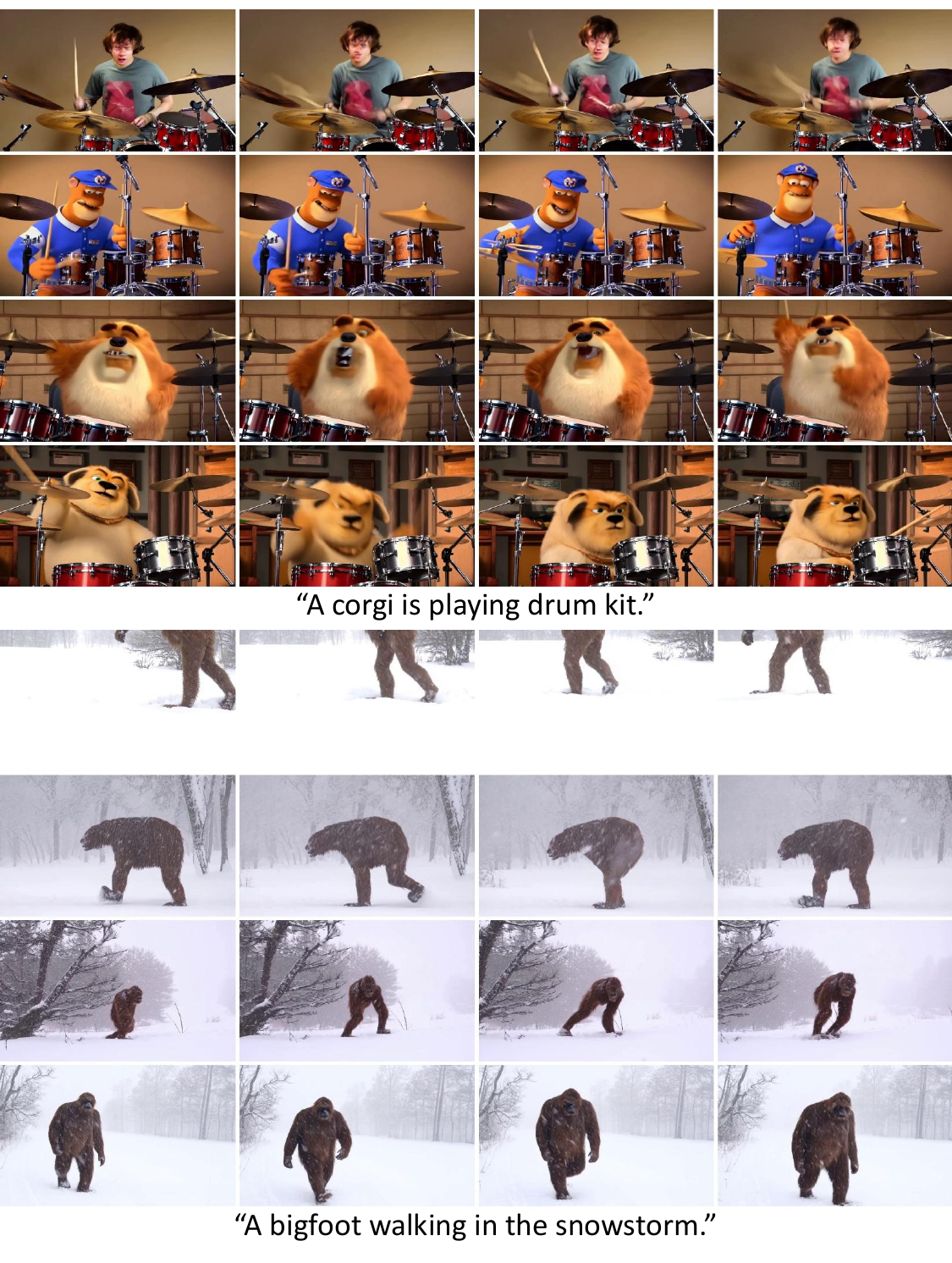}
    \caption{More visual results during TTS process on CogVideoX-5B.}
    \label{fig:visual_comparison_3}
\end{figure*}

\begin{figure*}[t]
    \centering
    \includegraphics[width=0.9\linewidth]{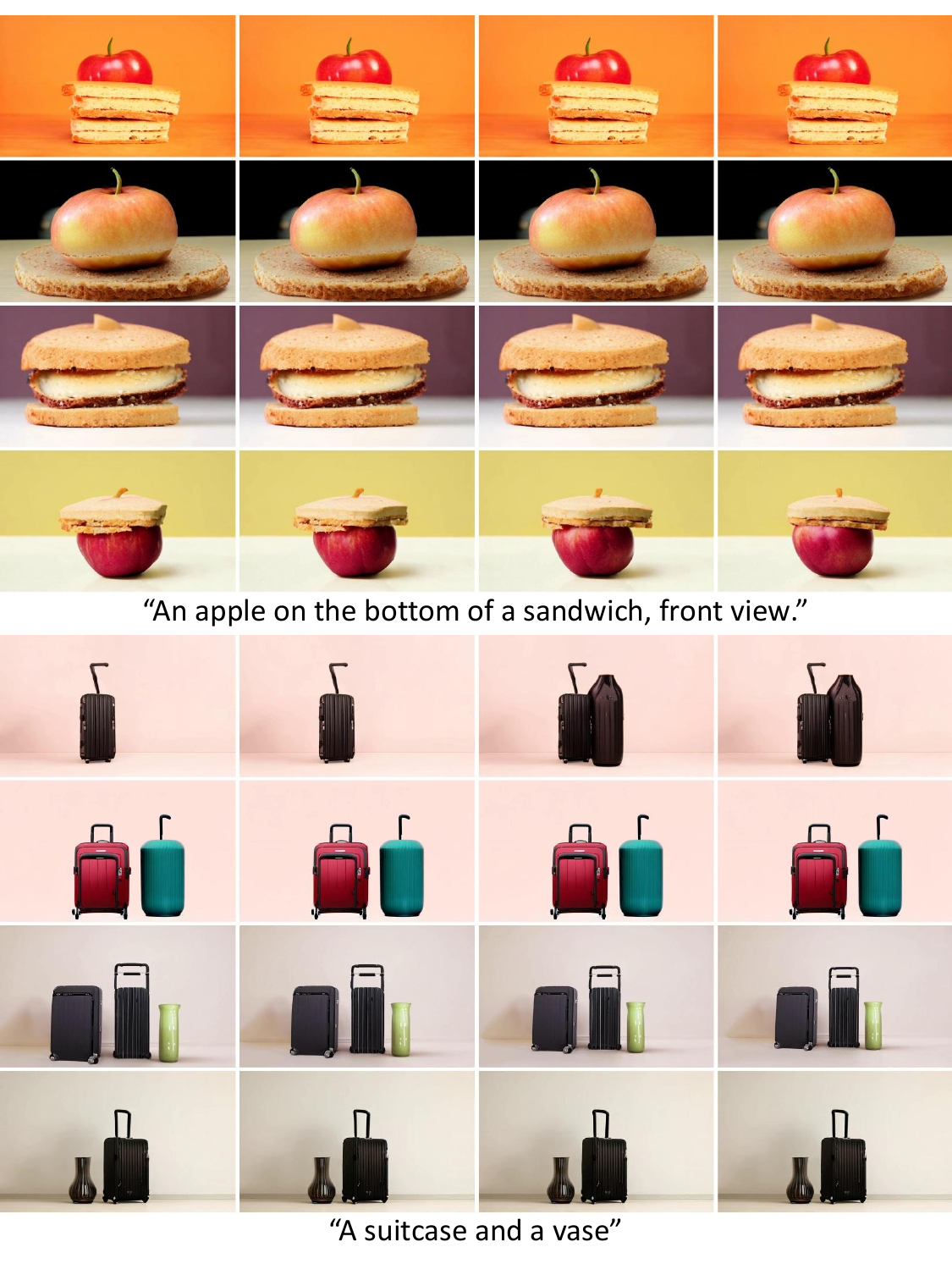}
    \caption{More visual results during TTS process on CogVideoX-2B. The TTS method can help to enhance multi-object spatial perception.}
    \label{fig:visual_comparison_4}
\end{figure*}

\begin{figure*}[t]
    \centering
    \includegraphics[width=0.92\linewidth]{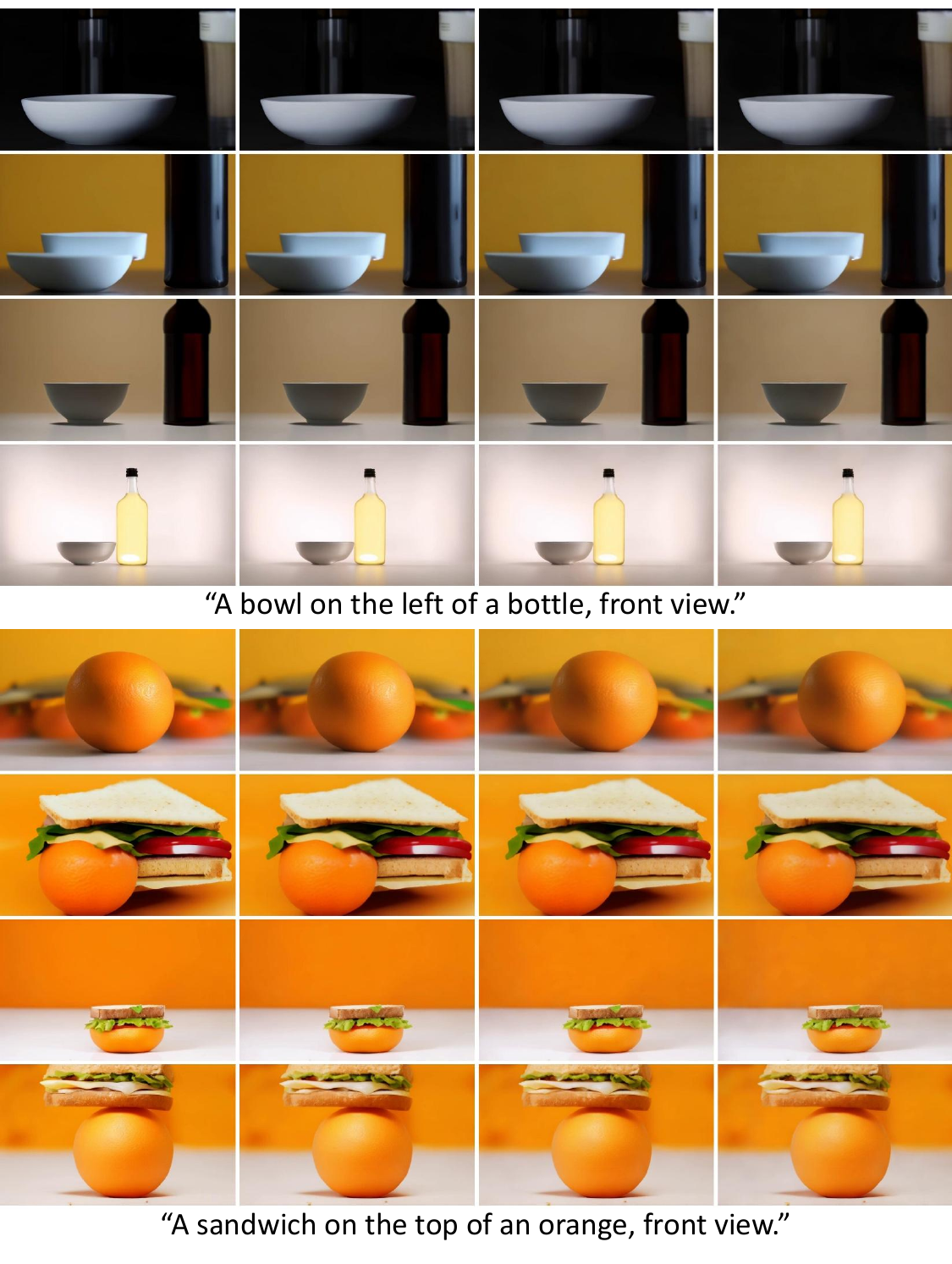}
    \caption{More visual results during TTS process on NOVA. The TTS method improves the alignment between the spatial relationships of objects in videos and the corresponding text prompt.}
    \label{fig:visual_comparison_5}
\end{figure*}

Figure~\ref{fig:failure_cases} shows failure cases in Pyramid-Flow experiments where TTS does not significantly improve video quality. Increasing inference computation fails to generate reasonable details (\eg, hand movements), indicating that model performance limitations constrain TTS method improvements.

\vspace{-5mm}
\vspace{-5mm}


\end{document}